\documentclass[journal]{IEEEtran}
\usepackage{graphicx}
\usepackage{caption}
\usepackage{algorithm}
\usepackage{algpseudocode}
\usepackage{booktabs}
\usepackage{multirow}
\usepackage{makecell}
\usepackage{cite}

\usepackage{mdframed}
\usepackage{xcolor}
\usepackage{colortbl}
\usepackage{listings}
\usepackage{enumitem}

\usepackage{balance}
\usepackage{soul}

\usepackage{amssymb}
\usepackage{amsmath}
\usepackage[
    colorlinks,
    linkcolor=red,
    anchorcolor=red,
    citecolor=green,
    urlcolor=magenta
]{hyperref}

\ifCLASSINFOpdf
\else
\fi
\begin{document}
%
\title{PosterLLaVa: Constructing a Unified Multi-modal Layout Generator with LLM}
%
%
%

\author{Tao~Yang$\dagger$,
        Yingmin~Luo$\dagger$,
        Zhongang~Qi$\ddagger$,
        Yang~Wu,~\IEEEmembership{Member,~IEEE,}
        Ying~Shan,
        and Chang~Wen~Chen$\ddagger$,~\IEEEmembership{Fellow,~IEEE}
\thanks{$\dagger$: Authors contributed equally to this research. $\ddagger$: Corresponding authors.}
\thanks{Tao~Yang, Changwen~Chen are with Hong Kong Polytechnic University.}
\thanks{Tao~Yang, Yingmin~Luo, Zhongang~Qi, and Ying~Shan are with Tencent PCG ARC Lab. Yang~Wu is with Tencent AI Lab.}
\thanks{Manuscript received April 19, 2005; revised August 26, 2015.}}

%
%

\markboth{Journal of \LaTeX\ Class Files,~Vol.~14, No.~8, August~2015}%
{Shell \MakeLowercase{\textit{et al.}}: Bare Demo of IEEEtran.cls for IEEE Journals}
%



\maketitle

\begin{abstract}
Layout generation is the keystone in achieving automated graphic design, requiring arranging the position and size of various multi-modal design elements in a visually pleasing and constraint-following manner. Previous approaches are either inefficient for large-scale applications or lack flexibility for varying design requirements. Our research introduces a unified framework for automated graphic layout generation, leveraging the multi-modal large language model (MLLM) to accommodate diverse design tasks.  In contrast, our data-driven method employs structured text (JSON format) and visual instruction tuning to generate layouts under specific visual and textual constraints, including user-defined natural language specifications. We conducted extensive experiments and achieved state-of-the-art (SOTA) performance on public multi-modal layout generation benchmarks, demonstrating the effectiveness of our method. Moreover, recognizing existing datasets' limitations in capturing the complexity of real-world graphic designs, we propose two new datasets for much more challenging tasks (user-constrained generation and complicated poster), further validating our model's utility in real-life settings. Marking by its superior accessibility and adaptability, this approach further automates large-scale graphic design tasks. Finally, we develop an automated text-to-poster system that generates editable SVG posters based on users' design intentions, bridging the gap between layout generation and real-world graphic design applications. This system integrates our proposed layout generation method as the core component, demonstrating its effectiveness in practical scenarios. The code and datasets are open-sourced on \url{https://github.com/posterllava/PosterLLaVA}.
\end{abstract}


%
\IEEEpeerreviewmaketitle

\section{Introduction}
%
%
%
%
\IEEEPARstart{F}{or} diverse sorts of graphic design (commercial posters, mobile app UIs, webpages, video thumbnails, etc.), layout plays a critical role in structuring visual and textual elements to captivate audiences and communicate intended messages. This task has required designers to create layouts manually, demanding their extensive expertise and experience. For large-scale designing tasks, the efficiency of this strategy is far from expected.

The most naive idea for massive graphic design generation is to utilize pre-design templates and replace content according to requirements. However, the production and selection of templates still involve human labor, and mechanically applying the inappropriate layout template can lead to obtrusive designs. Previous researchers attempted to frame layout generation as an optimization problem, tackling it with heuristic algorithms~\cite{rule_based_layout0, rule_based_layout1, rule_based_layout2}. However, these methods hinge on crafting well-designed energy functions, a task that still depends heavily on design expertise, and the form of which usually lacks generality across different applications.

\begin{figure}[t]
  \includegraphics[width=0.99\linewidth]{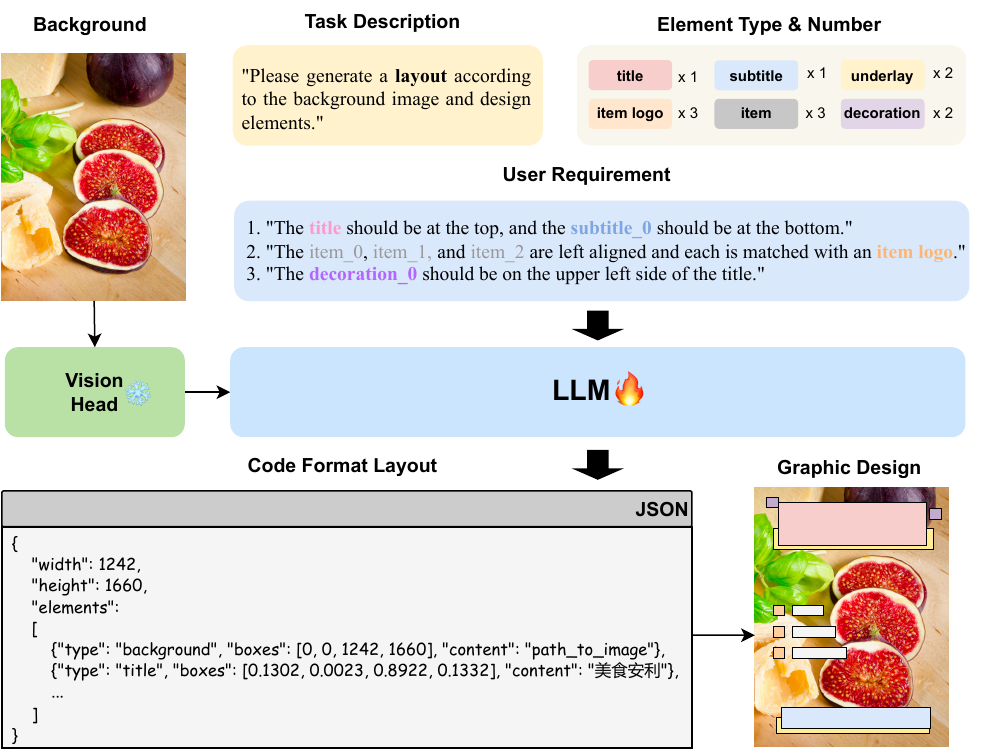}
  \caption{The overall framework of our proposed content-aware layout generation method. Adopting the multi-modal LLM \cite{LLaVa} as the central processing unit, we embed information from visual and textual domains to generate a reasonable and visually pleasing graphic layout. The result is encoded in JSON format and can be rendered into a real-world poster.}
  \vspace{-10pt}
  \label{fig_overall_framework}
\end{figure}

With the advance in deep learning, researchers are glad to embrace data-driven methods \cite{LayoutVAE, LayoutGANpp, LayoutTransformer, VTN, BLT, LayoutDM} in layout generation. Most of these works focus on adopting the latest generative architecture but overlook the necessary conditional requirements for layout. This limits their applicability in real-world scenarios, which frequently demand the integration of complex multi-modal conditions. Recently, more and more researchers have recognized the importance of multi-modal conditions and started to explore \textit{content-aware layout generation}. For visual conditions, CGL-GAN \cite{CGL_GAN} and DS-GAN \cite{PosterLayout} take an innovative step to incorporate the semantic information on background images as conditions in layout generation, and some later work \cite{LayoutDETR, HPCVTG} also consider the content of foreground elements as conditions. For textual conditions, some preliminary attempts \cite{LayoutGANpp, LayoutFormerpp, Parse_then_Place} generate layouts under given graphic conditions. However, the introduced constrained optimization processes or specific intermediate representations strengthen the training or labeling complexity. An efficient end-to-end framework that can directly translate natural language instructions into desired layouts is still needed.

Although previous approaches have demonstrated progress on certain datasets, most of them rely on highly customized network architectures that lack universality. Such specificity necessitates substantial modifications or complete redesigns to accommodate new or varied layout design challenges. Recognizing this limitation, we develop a unified framework named PosterLLaVa (see Fig. \ref{fig_overall_framework}) for layout generation task, inspired by the simplicity and effectiveness of the recently published multi-modal instruction tuning \cite{BLIP2, miniGPT4, LLaVa, llama_adapter, mplug_owl, VisionLLM} method. Pre-trained with numerous amounts of unlabelled corpora and fine-tuned with instruction-following data, MLLMs (Multi-modal Large Language Models) are capable of handling multiple vision-language tasks (e.g., VQA \cite{miniGPT4, LLaVa}, visual grounding \cite{VisionLLM, mplug_owl}, etc.) according to the given instructions and their knowledge. 
For layout generation, we first show how layout information can be represented by structure natural language in JSON format.
With this representation, we can measure the performance of PosterLLaVa on established content-aware generation datasets and compare it with previous benchmarks. 
To tackle the multi-modal condition inputs, we utilize the pre-trained visual head of LLaVa \cite{LLaVa_1d5} to convert input image to representation adapted to textual token space and fine-tuning the LLM \cite{LLaMa2} to interpret and generate layout data. 
With the LLM as the central processing unit, our model can manage a wide range of layout generation tasks through simple modifications of the input instructions, eliminating any need for changes in model architecture. Moreover, user requirements presented by natural language can be seamlessly integrated into the generation instructions, enhancing the model's responsiveness to specific design needs.

The main contribution of our work can be summarized as follows. 

\begin{enumerate}
    \item \textbf{A Unified Layout Generation Tool} We propose a unified content-aware layout generation method using multi-modal LLMs, adaptable across various design scenarios through simple modifications of input instructions. Our approach is validated across multiple public datasets and two new datasets proposed in this paper, showcasing its superior performance and versatility.
    \item \textbf{New Datasets for Real-world Applications} Recognizing the shortcomings of existing content-aware layout generation datasets in handling real-world demands, we collect data for two challenging tasks. The first is a graphic layout dataset named QB-Poster, composed of 5,188 samples designed with prevalence on Chinese social networks. This dataset is characterized by its intricate geometric relationships between sufficient types of content. The second is the UC-Poster, which is the largest poster layout dataset composed of 64,520 layouts, each constrained by both a background image and 3 textual user requirements. Through comparative analysis with the latest comparable method, our method demonstrates remarkable adaptability and effectiveness in capturing the distribution of complicated real-world layouts.
    \item \textbf{An Automatic Text-to-editable-poster System} With PotserLLaVa as the fundamental component, we build a novel pipeline named PosterGen to interpret the user's intention and then create a visually appealing poster design to convey the key information. PosterGen utilizes the latest LLMs and T2I diffusion models. We provide experiments to prove its production quality.
\end{enumerate}

\section{Related Work}
\subsection{Automatic Graphic Layout Generation} \label{related_work}

\textbf{Rule-based Methods} Before the appearance of deep learning, layout generation has been studied for decades~\cite{rule_based_layout0, rule_based_layout1, rule_based_snippet, rule_based_magazine}. Typically, Yin et al.~\cite{rule_based_snippet} proposed a series of principles according to widely accepted aesthetic or information-conveying rules and a heuristic algorithm to minimize the overall energy function. These methods do not require training. Instead, they perform a runtime searching process during every inference. The true complexity of these methods lies in the design of the energy function, which requires a lot of design experience and expertise. Moreover, these functions must be manually re-designed when encountering a new design element or applied to a different styled layout (e.g., from UI to commercial poster). 

\textbf{Content-agnostic Layout Generation} Neural networks offer researchers a way to formulate designing principles implicitly from numerous data, saving human efforts. Most early works \cite{LayoutGAN, LayoutVAE, ContentGAN, LayoutTransformer, VTN, LayoutDiffusion} focus on generating visually reasonable layouts for mobile UIs, documents, and magazine pages. LayoutGAN \cite{LayoutGAN} employs the GAN (Generative Adversarial Network) paradigm and designs a differentiable rendering process for connecting the visual and graphic domains. LayoutVAE \cite{LayoutVAE} and CanvasVAE\cite{CanvasVAE} adopt the VAE (Variational Auto-Encoder) paradigm, while more recent works adopt the auto-regressive architecture \cite{LayoutTransformer, VTN, BLT} or the diffusion architecture~\cite{LayoutDM, LayoutDiffusion, LDGM}. Despite their achievement on unconditioned layout generation tasks, they are hard to use in real-world scenarios.

\textbf{Content-aware Layout Generation} Recently, some other works \cite{PosterLayout, CGL_GAN, LayoutDETR, PDA_GAN, HPCVTG} have paid their attention to commercial-style posters, in which case the graphic designs are usually based on a non-empty background image. CGL-GAN~\cite{CGL_GAN} contributes a large dataset with around 60k Chinese commercial posters and proposes to learn with a transformer-based GAN network receiving a saliency map and the inpainted background as input. Similarly, PosterLayout~\cite{PosterLayout} tackles the problem with a CNN-LSTM network with saliency map as input. \cite{ICVT} adopts a C-VAE (Conditional Variational Auto-Encoder) to predict the layout. LayoutDETR~\cite{LayoutDETR} design a DETR-like\cite{DETR} to utilize the pre-trained objects detection model and integrate both GAN and VAE for layout generation. They also include pre-trained ViT \cite{ViT} and BETR \cite{BERT} as visual and textual encoders to get embedded features of the design elements. Interestingly, some work \cite{LayoutGANpp, LayoutFormerpp, Parse_then_Place} also attempted to generate layouts following specific constraints. Primitively, LayoutGAN++~\cite{LayoutGANpp} introduces an additional constrained optimization process based on the Lagrangian multiplier method to get the desired layout. Then, LayoutFormer++~\cite{LayoutFormerpp} and Parse-then-place~\cite{Parse_then_Place} design a specific intermediate representation to handle various constraints. The latter also studies the text-to-layout problem, which includes implicitly expressed user requirements and is very similar to ours.

\subsection{Multi-modal Large Language Models and Application}
\textbf{LLMs (Large Language Models)} \cite{GPT3, GPT4, LLaMa2} have achieved remarkable success across a wide range of natural language processing (NLP) tasks. With billions of parameters, these models derive extensive knowledge from pre-training on vast unlabeled text corpora. Various instruction-tuning methods have been investigated to enhance the ability of LLMs to comprehend and execute natural language instructions \cite{ouyang2022training, wang2022self}. While LLMs have proven adept at understanding and generating text, multi-modal LLMs have been facilitated by incorporating additional modalities like visual and auditory data \cite{BLIP2, miniGPT4, LLaVa}. A prevalent approach involves injecting LLMs with multi-modal information and leveraging their robust reasoning capabilities.

\textbf{LLMs-assisted Layout Generation} Layouts, which can be encoded in formats such as XML or JSON, are ideally suited to be processed by pre-trained Large Language Models (LLMs). Previous works have used domain-specific data to strengthen their code generation ability. LayoutNUWA \cite{LayoutNUWA} fine-tunes the LLaMa \cite{LLaMa} and CodeLLaMa \cite{CodeLLaMa} to the content-agnostic layout generation task, achieving the SOTA performance in multiple content-agnostic layout datasets. LayoutPrompter \cite{LayoutPrompter} introduces an interesting training-free approach, leveraging RAG (Retrieval-Augmented Generation) to strengthen the in-context learning ability of GPT \cite{GPT3}, dynamically sourcing examples from a dataset. However, this retrieval-centric strategy is limited to open-domain generation. These works overlook the visual domain feature or translate it into hard tokens before feeding into LLM, potentially resulting in severe information loss. To tackle this weakness, we include the latest proposed multi-modal technique - visual instruct tuning \cite{LLaVa} to fine-tune a pre-trained large model, which accepts the visual information with a pre-trained and aligned visual adaptation head \cite{CLIP}. For the layout-to-image generation, interestingly, some contemporaneous work like LayoutGPT\cite{LayoutGPT} and TextDiffuser-2\cite{TextDiffuser2} also adopt LLMs, showing a promising production pipeline for LLM-based graphic design.

\subsection{Graphic Design Generation}
Most of the papers \cite{LayoutVAE, LayoutGAN, CGL_GAN, PosterLayout} mentioned previously have conducted research solely on layout generation, but the subsequent task \textbf{i.e., from a layout to a complete graphic design} still requires human effort. This includes theme-related images, copywriting, and font design. Considering this purpose, some early works \cite{Text2poster, HPCVTG} proposed naive schemes. Given a text prompt, Text2poster \cite{Text2poster} advises retrieving the background image with similar semantics and rendering the text element on it with the LSTM-generated layout and predefined font types, sizes, and colors. HPCVTG \cite{HPCVTG} develops a thumbnail generation system, given a video and its text description, it retrieves the frame with similar semantic meanings and renders the GPT-summarized text element with predefined font types and colors, while layouts are optimized with the EA (Evolution Algorithm). Recently, glyph-conditioned diffusion models \cite{GlyphDraw, GlyphControl, TextDiffuser, AnyText, Glyph-ByT5} have been utilized for scene-text generation or editing, showcasing a surprisingly better performance than vanilla stable diffusion. Nonetheless, most publicly available models face trouble when tackling text with numerous characters and small font sizes.

COLE \cite{cole} is the latest published work on generating whole-stage graphic layouts, which builds a system with diffusion model \cite{sd} and LLM layout planner \cite{LLaMa2}. However, it relies on a closed-sourced graphic design dataset, which requires large human effort to construct. The unavailability of the dataset and code has restricted the re-productivity of this well-designed method. For public availability considerations, OpenCOLE \cite{opencole} reproduced and simplified COLE's pipeline with a publicly available dataset (i.e. Crello \cite{CanvasVAE}), offering an open-sourced alternative with compatible performance. Nevertheless, it suggests fine-tuning both the diffusion model and topography-LLM, still consumes considerable computational resources. 

\begin{table*}[t]
    \centering
    \begin{tabular}{p{0.95\textwidth}}
    \toprule
    \textsc{\textbf{User:}} \\
    \textsf{\text{$<$image$>$}}\\
    Please help me to place \textsf{\text{$<$N$>$}} foreground elements over the background of \textsf{\text{$<$resolution$>$}} to craft a \textsf{\text{$<$domain\_name$>$}}. Remember to avoid unbalance, overlap, misalignment, and occlusion of semantic-meaningful objects on the background image. Return the result by filling in the following JSON file while keeping the number and types of elements unchanged. The initial JSON is defined as: \textsf{\text{$<$masked\_json$>$}}, in which each design element is represented by a bounding box described as [left, top, right, bottom], and each coordinate is a contiguous number in 0-1. The user constraints are defined as: \textsf{\text{$<$constraints$>$}}, which should be adopted as compulsory design requirements. \\
    \ \\
    \textsc{\textbf{Assistant:}} \\
    \text{Sure! Here is the design result:} \textsf{\text{$<$json$>$}}. \\
    \bottomrule
    \end{tabular}
    \caption{Prompt template for applying visual instruction tuning on content-aware generation task. The placeholder tokens in bold type are replaced with specific information during training or inference.} \label{prompt_template}
\end{table*}

\section{Methodology} \label{sec_method}

\subsection{Multi-modal Layout Tokenization} \label{section_text_layout}
Assuming that all complicated attributes and art styles have their default values, we can explicitly represent the information of a graphic design $L_{j}$ by defining the position $(x_{i}, y_{i})$, size $(h_{i}, w_{i})$, and content $\mathbf{I_{i}}$ of every element. The position and size can be further expressed as bounding box format if rotation and irregular shapes are not involved. The class labels $c_{i}$ of elements are explicitly given to excavate the relationship between different kinds of elements. We got the following representation of a poster:
\begin{equation}
\text{L}_{j} = \{(x_{i}, y_{i}, h_{i}, w_{i}), c_{i}, \mathbf{I}_{i}\}_{i=0}^{N}
\end{equation} 
in which $N$ represents the number of elements. For previous papers, most consider $L_{j}$ as a numeric form, which means solving the problem in a continuous space. 
However, we design the following process to tokenize $L_{j}$ and feed it into LLMs to predict the next token. First, we normalized the bounding box coordinates with the background width and height to facilitate multi-resolution generation. 
Each coordinate data value of the bounding box vector is truncated to $K$ decimal places to avoid redundancy. For class label $c_{i}$, we use the corresponding text label instead, for example $\{\text{text}, \text{logo}, \text{underlay}\}$ regarding the PosterLayout~\cite{PosterLayout} dataset. Finally, for image elements, $\mathbf{I^{\text{img}}_i}$ is encoded by a pre-trained vision header, which is composed of a ViT~\cite{ViT} encoder and a linear projection head, namely
\begin{equation} \label{eq_llava}
    h(\mathbf{I}_{i}^{\text{img}}) = \mathbf{W}^T \text{CLIP}(\mathbf{I}_{i}^{\text{img}}).
\end{equation} 
and the content $\mathbf{I^{\text{txt}}}$ of the text element is inherently in a text format.


\subsection{Training Scheme}
To facilitate the learning of tokenized layout data, we adopt the training scheme proposed by Liu et al. \cite{LLaVa}, i.e., the visual instruction tuning. The original paper, focusing on general visual-language tasks, recommends fine-tuning a pre-trained LLM \cite{LLaMa} by two phrases: 1. pre-training for feature alignment, and 2. end-to-end fine-tuning. The alignment phase usually requires numerous image-text pairs to adapt visual information into language space, and the fine-tuning phase requires relatively less data to acquire instruction-following outputs. Recognizing that the primary challenge in layout generation resides in decoding the semantic and geometric relationship between graphic elements, we streamline the training process by using the pre-trained linear projection layer to skip the feature alignment phase. This allows us to reduce training expenditure while maintaining comparable performance with the full-trained model.

\subsection{Prompt Template}
We introduce the following prompt template for the adopting end-to-end fine-tuning phase of visual instruction tuning in various content-aware layout generation tasks. The template is described in Tab.~\ref{prompt_template}. The pre-trained vision head converts the background image into soft tokens (as Eq.~\ref{eq_llava} shows) to get \textsf{\text{$<$image$>$}}. \textsf{\text{$<$N$>$}} is replaced with the exact number of design elements, and \textsf{\text{$<$resolution$>$}} is replaced with the canvas resolution. We use a domain indicator \textsf{\text{$<$domain\_name$>$}} to distinguish different tasks and datasets. For example, "commercial poster" for CGL dataset and "advertising banner" for ad banner dataset. The ground-truth layout information is expressed by textual representation through the process introduced in Sec.~\ref{section_text_layout} and arranged in JSON format (as Fig.~\ref{fig_overall_framework}) to replace \textsf{\text{$<$json$>$}}. For human instruction, we delete bounding boxes and preserve the category labels to get the \textsf{\text{$<$masked\_json$>$}}. As for user-constrained generation tasks, the constraints are given as \textsf{\text{$<$constraints$>$}}.

\section{Experiment}

\textbf{Implementation Details} Most experiments are conducted on 8 NVIDIA A10 GPUs and can be finished within 12 hours. The MLLM checkpoint adopted is the full-tuning 7B version of LLaVa-v1.5 \cite{LLaVa_1d5}, which is trained with LLaMa-2 \cite{LLaMa2} 7B as the base model with visual instruction tuning. For most of the following layout datasets, we fine-tune the MLLM with one epoch. But for the ad banners dataset, considering its tiny scale, we find that the model needs at least 3 epochs to converge. For the adaptation into the QB-Poster dataset, we adopt the pre-trained model on all training sets of Ad Banner, CGL, and PosterLayout as a starting point to enhance its performance. We increase the max token from 2048 to 4096 as the token length grows with the element number. For other training or inference hyper-parameters, we apply the default recipe recommended by LLaVa \cite{LLaVa}. 

\subsection{Result on Public Content-aware Layout Dataset}

\textbf{Dataset Description} As mentioned in Section~\ref{related_work}, content-aware layout generation, 
is still in its early stages. For learning this task, a corresponding background image and the coordinates of  ground-truth layout bounding boxes are needed. We extensively investigated datasets published in previous literature.
Available public datasets are listed in Tab.~\ref{tab_dataset}.

\begin{table}[t]
    \centering
    \setlength{\tabcolsep}{5pt}
    \caption{An overall description of the content-aware layout generation datasets. QB-Poster is the complicated real-world poster dataset proposed in this paper, which outperforms previous datasets in both annotation categories and box numbers per poster.}
    \label{tab_dataset}
    \begin{tabular}{ccccccc}
        \toprule
        \textbf{Dataset} & \textbf{Train} & \textbf{Test}  & \textbf{Classes} & \textbf{Boxes/img} & \textbf{Total Boxes} \\
        \midrule
        CGL dataset                 & 60,548 & 1,000  & 4 & 4.87 & 265,818 \\
        PosterLayout                & 9,974 & 905  & 3 & 4.73 & 47,024 \\
        Ad Banner                   & 7,672 & 1,000  & 8 & 2.23 & 16,593 \\
        YouTube                     & 10,000 & 1,000       & 3 & 5.88 & 67,223 \\
        \rowcolor{gray!25}
        QB-Poster                   & 4,675 & 513  & \textbf{10} & \textbf{15.17} & \underline{78,723} \\
        \bottomrule
    \end{tabular}
\end{table}

CGL dataset~\cite{CGL_GAN}, one of the pioneering content-aware poster layout collections, comprises 60,548 training samples and 1,000 test samples collected from e-commerce platforms. The design elements are divided into 4 categories: logo, text, underlay, and embellishment. The class labels and the bounding boxes of the elements for each poster in the training set are manually annotated, while the test set only includes the background image. Techniques such as image inpainting~\cite{LaMa} and saliency detection~\cite{PFPN} are needed to obtain additional visual information. Recognizing the limitations of the CGL dataset, particularly its repetitive content and scarcity of complex layouts featuring over ten elements, Hsu et al.~\cite{PosterLayout} introduce PosterLayout, offering 9,974 poster-layout pairs for training and 905 background images for testing. LayoutDETR~\cite{LayoutDETR} contributes an ad banner dataset with multi-modal information, containing 7,672 samples divided into training and testing subsets in a 9: 1 ratio. The background images are either from the Pitt Image Ads dataset or Google Image, and the bounding boxes, categories, and text content are automatically extracted by OCR automatically. However, unlike CGL and PosterLayout, this dataset contains banners with multi-resolutions. The YouTube~\cite{HPCVTG} dataset is another newly proposed dataset focusing on video thumbnail generation. The design elements are images and texts extracted from the original video, and the ground-truth layouts are generated using an evolution algorithm with hand-crafted design principles as objectives.

\begin{table}[t] 
    \centering
    \begin{minipage}[t]{0.48\textwidth} 
        \setlength{\tabcolsep}{1.5pt}
        \scriptsize
        \caption{Results comparison on PosterLayout dataset. Evaluations are conducted under PosterLayout's~\cite{PosterLayout} settings. Previous results are copied for comparison.} \label{result_posterlayout}
        \begin{tabular}{lcccccccc}
            \toprule
            \multirow{2}[4]{*}{\makecell{\textbf{Methods}}} &  \multicolumn{3}{c}{\textbf{Content-aware}} & \multicolumn{5}{c}{\textbf{Geometric}} \\
            \cmidrule(lr){2-4} \cmidrule(lr){5-9}
            & \makecell{Uti$\Uparrow$} & \makecell{Occ$\Downarrow$} & \makecell{Rea$\Downarrow$} & \makecell{Val $\Uparrow$} & \makecell{Ove $\Downarrow$} & \makecell{Ali $\Downarrow$} & \makecell{$\text{Und}_{l}\Uparrow$} & \makecell{$\text{Und}_{s}\Uparrow$} \\ 
            \midrule
            Ground-Truth & 0.2222 & 0.1900 & 0.1522 & 0.9999 & 0.0001 & 0.0002 & 0.9965 & 0.9912 \\
            \midrule
            & \multicolumn{8}{c}{\textbf{Content-aware Methods}} \\
            CGL-GAN & 0.2257 & 0.1546 & 0.1715 & 0.7066 & 0.0605 & 0.0062 & 0.8624 & 0.4043 \\
            DS-GAN\cite{PosterLayout} & 0.2541 & 0.2088 & 0.1874 & 0.8788 & 0.0220 & 0.0046 & 0.8315 & 0.4320 \\
            LayoutPrompter\cite{LayoutPrompter} & 0.2597 & \textbf{0.0992} & 0.1723 & 0.9992 & 0.0036 & 0.0036 & 0.8986 & 0.8802 \\
            \rowcolor{gray!25}
            \textbf{PosterLLaVa(Ours)} & \textbf{0.2628} & 0.1649 & \textbf{0.1142} & \textbf{1.0000} & \textbf{7.7e-5} & \textbf{0.0002} & \textbf{1.0000} & \textbf{1.0000} \\
            \bottomrule
        \end{tabular}
    \end{minipage}
    \begin{minipage}[t]{0.48\textwidth}
        \centering
        \footnotesize
        \setlength{\tabcolsep}{0.4pt}
        \vspace{8pt}
        \caption{Results comparison on CGL-GAN dataset. Evaluations are conducted under CGL-GAN's~\cite{CGL_GAN} settings. Previous results are copied for comparison. $\dagger$ indicates that we apply BASNet~\cite{BASNet} for saliency detection rather than PFPN~\cite{PFPN} since the pre-trained link of the latter one expires.} \label{result_cgl}
        \begin{tabular}{lccccccc}
            \toprule
            \multirow{2}[4]{*}{\makecell{\textbf{Methods}}} &  \multicolumn{3}{c}{\textbf{Content-aware}} & \multicolumn{4}{c}{\textbf{Geometric}} \\
            \cmidrule(lr){2-4} \cmidrule(lr){5-8}
            & \makecell{$R_{\text{com}}$ $\Downarrow$} & \makecell{$R_{\text{shm}}$ $\Downarrow$} & \makecell{$R_{\text{sub}}$ $\Downarrow$} & \makecell{$R_{\text{ove}}$ $\Uparrow$} & \makecell{$R_{\text{und}}$ $\Uparrow$} & \makecell{$R_{\text{ali}}$ $\Uparrow$} & \makecell{$R_{\text{occ}}$ $\Uparrow$} \\
            \midrule
            & \multicolumn{7}{c}{\textbf{Content-unaware Methods}} \\
            LayoutTransformer~\cite{LayoutTransformer} & 40.92 & 21.08 & 1.310 & 0.0156 & 0.9516 & 0.0049 & - \\
            VTN~\cite{VTN} & 41.77 & 22.21 & 1.323 & 0.0130 & 0.9628 & 0.0047 & - \\
            \midrule
            & \multicolumn{7}{c}{\textbf{Content-aware Methods}} \\
            ContentGAN~\cite{ContentGAN} & 45.59 & 17.08 & 1.143 & 0.0397 & 0.8626 & 0.0071 & 93.4 \\
            CGL-GAN~\cite{CGL_GAN} & 35.77 & 15.47 & 0.805 & 0.0233 & 0.9359 & 0.0098 & 99.6 \\
            PDA-GAN~\cite{PDA_GAN} & \textbf{33.55} & 12.77 & 0.688 & 0.0290 & 0.9481 & 0.0105 & 99.7 \\
            \rowcolor{gray!25}
            \textbf{PosterLLaVa(Ours)} & \underline{34.80} & \textbf{8.214} & \textbf{0.277}$\dagger$ & \textbf{2.4e-10} & \textbf{1.0000} & \textbf{0.0008} & \textbf{100} \\
            \bottomrule
        \end{tabular}
    \end{minipage}
    \vspace{-10pt}
\end{table}

\textbf{Evluation Metrics} 
For a fair comparison with results published in previous papers, we first adopt the original evaluation measurements for each dataset. The metrics used are similar for CGL-dataset \cite{CGL_GAN} and PosterLayout \cite{PosterLayout} dataset. The calculation of content-aware metrics is related to background or saliency image: the $R_{\text{com}}$ and Rea represent the readability of text elements; $R_{\text{shm}}$, $R_{\text{sub}}$, Occ represents the occlusion of semantic meaningful or saliency region on the background, while Uti indicates the utility of non-saliency region. The geometric metrics are only related to the predicted bounding boxes: $R_{\text{ove}}$, Ove represents the overlap ratio; $R_{\text{und}}$, $\text{Und}_{l}$ and $\text{Und}_{s}$ indicates whether the underlays are correctly placed under texts, and Ali represents the alignment; $R_{\text{occ}}$ and Val indicates the valid (e.g., non-empty) layout ratio. For Ad Banner \cite{LayoutDETR} and YouTube \cite{HPCVTG} dataset, similarity metrics are included since the ground-truth layouts are available. This is achieved by measuring the FID (Frechet Inception Distance) or IoU between the generated layout/image and the corresponding ground-truth.  VB in the YouTube dataset represents Visual Balance, which represents whether the overall placement is balanced. Please refer to the original papers for detailed explanations of these metrics.

\begin{table}[t]
    \centering
    \begin{minipage}[t]{0.48\textwidth} 
        \footnotesize
        \centering
        \caption{Results comparison on the ad banner dataset under LayoutDETR's~\cite{LayoutDETR} settings. Results of previous methods are copied for comparison, among which PosterLLaVa achieves SOTA performance in all metrics except misalignment.} \label{result_banners}
        \setlength\tabcolsep{1pt}
        \begin{tabular}{lcccccc} 
            \toprule
            \multirow{2}[4]{*}{\makecell{\textbf{Methods}}} & \multicolumn{4}{c}{\textbf{Similarity}} & \multicolumn{2}{c}{\textbf{Geometric}} \\
            \cmidrule(lr){2-5} \cmidrule(lr){6-7}
            & \makecell{Layout\\FID$\Downarrow$} & \makecell{Image\\FID$\Downarrow$} & \makecell{IoU\\$\Uparrow$} & \makecell{DocSim\\$\Uparrow$} & \makecell{Overlap\\$\Downarrow$} & \makecell{Misalign\\$(\times10^{-2})\Downarrow$} \\
            \midrule
            Ground-Truth & - & - & - & - & 0.035 & 1.889 \\
            \midrule
            & \multicolumn{6}{c}{\textbf{Content-unaware Methods}} \\
            LayoutGAN++ \cite{LayoutGANpp} & 4.25 & 28.40 & 0.163 & 0.130 & 0.104 & 0.759 \\
            READ  & 4.45 & 32.10 & 0.177 & 0.141 & 0.093 & 2.867 \\
            Vinci & 38.97 & 58.12 & 0.104 & 0.143 & 0.243 & \textbf{0.271} \\
            LayoutTransformer \cite{LayoutTransformer} & 5.47 & 39.70 & 0.080 & 0.115 & 0.127 & 3.632 \\
            \midrule
            & \multicolumn{6}{c}{\textbf{Content-aware Methods}} \\
            CGL-GAN \cite{CGL_GAN} & 4.69 & 30.50 & 0.154 & 0.127 & 0.116 & 1.191 \\
            ICVT \cite{ICVT}  & 12.54 & 30.11 & 0.163 & 0.137 & 0.423 & 0.682 \\
            LayoutDETR-VAE \cite{LayoutDETR} & 3.25 & 27.47 & 0.216 & 0.152 & 0.119 & 1.737 \\
            \rowcolor{gray!25}
            \textbf{PosterLLaVa(Ours)} & \textbf{2.37} & \textbf{24.87} & \textbf{0.242} & \textbf{0.158} & \textbf{0.029} & 1.161 \\
            \bottomrule
        \end{tabular}
    \end{minipage}
    \begin{minipage}[t]{0.48\textwidth} 
        \footnotesize
        \centering
        \vspace{8pt}
        \caption{Results comparison on the Youtube dataset under HPCVTG's \cite{HPCVTG} settings. Previous results are copied for comparison. PosterLLaVa shows promising performance in reducing overlap and saliency occlusion.
        } \label{result_youtube}
        \setlength\tabcolsep{1.8pt}
        \begin{tabular}{lcccccc}
            \toprule
            \multirow{3}{*}{\makecell{\textbf{Methods}}} & \multicolumn{2}{c}{\textbf{Similarity}} & \multicolumn{4}{c}{\textbf{Geometric}} \\
            \cmidrule(lr){2-3} \cmidrule(lr){4-7} 
            & \makecell{mIoU\\$\Uparrow$}  & \makecell{FID\\$\Downarrow$} & \makecell{VB\\$\Downarrow$} & \makecell{Overlap\\$\Downarrow$} & \makecell{Misalign\\$\Downarrow$} & \makecell{Occlusion\\$\Downarrow$} \\
            \midrule
            Ground-Truth & - & - & 0.93 & 6.29 & 1.55 & 5.88 \\
            \midrule
            & \multicolumn{6}{c}{\textbf{Content-unaware Methods}} \\
            LayoutGAN++ \cite{LayoutGANpp} & 4.06 & 145.7 & 6.01 & 151.02 & 1.52 & 21.23 \\
            LayoutTransformer \cite{LayoutTransformer} & 11.42 & 59.89 & 6.53 & 76.15 & \textbf{0.06} & 18.38 \\ 
            \midrule
            & \multicolumn{6}{c}{\textbf{Content-aware Methods}} \\
            HPCVTG \cite{HPCVTG} & 14.16 & 18.50 & \textbf{2.13} & 47.51 & 3.25 & 14.41 \\
            \rowcolor{gray!25}
            \textbf{PosterLLaVa(Ours)} & \textbf{27.50} & \textbf{12.14} & \underline{3.10} & \textbf{8.17} & \underline{0.49} & \textbf{7.24} \\
            \bottomrule
        \end{tabular}
    \end{minipage}
    \vspace{-10pt}
\end{table}

\textbf{Result Comparison} The results presented in Tab. \ref{result_posterlayout}, \ref{result_cgl}, \ref{result_banners}, and \ref{result_youtube} demonstrate that our method outperforms existing approaches, both content-unaware and content-aware, by a significant margin. In the Ad Banner dataset, our model exhibits improvements across all metrics except Misalign. For the PosterLayout dataset, our method markedly enhances geometric metrics, whereas LayoutPrompter~\cite{LayoutPrompter} achieves a better trade-off between utility and occlusion. This is understandable because all previous methods incorporate additional input (i.e., saliency maps pre-processed by the saliency detector), while our method relies solely on the original background image. Similarly, in the CGL dataset, our method outperforms other approaches, particularly in geometric measurements. These results confirm the effectiveness of our method across various datasets and metrics.

\begin{figure*}[t]
    \centering
    \includegraphics[width=0.95\linewidth]{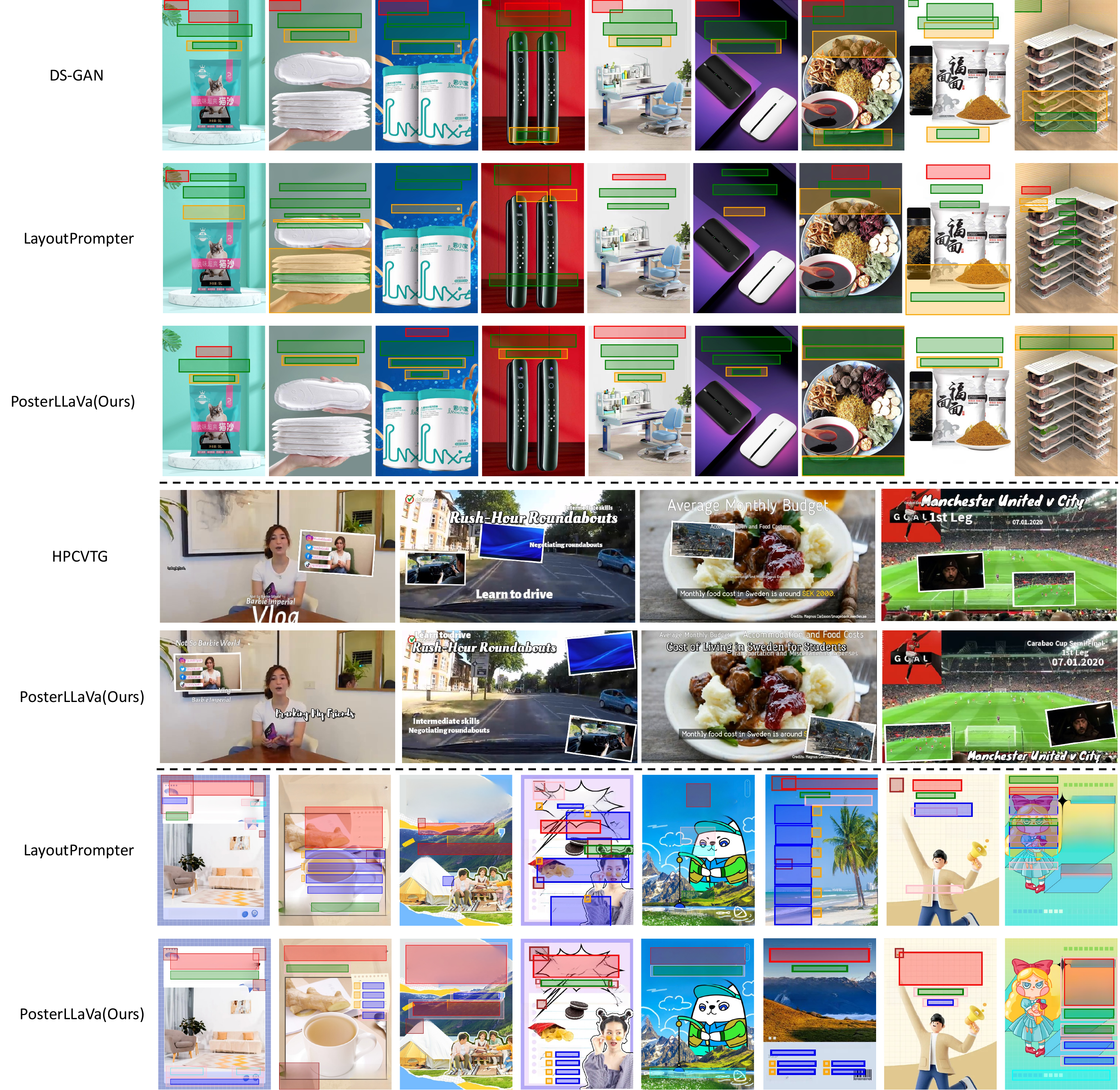}
    \caption{Qualitative results on the PosterLayout (top), Youtube (middle), and QB-Poster (bottom) datasets. PosterLLaVa achieves the highest overall generation quality on all three datasets.}
\end{figure*}

\subsection{Result on the Proposed Datasets} \label{subsection_new_datasets}


\textbf{User-constrained Layout Generation} Although content-aware layout generation has been a valuable step toward real-world applications, realistic graphic design problems often involve more conditionality. User constraint is one of them, usually including optional suggestions or mandatory opinions for graphic design products. These constraints, typically articulated in natural language, introduce even more complexity due to their potential ambiguity. As Section \ref{related_work} mentioned, several previous works \cite{LayoutGANpp, LayoutFormerpp, Parse_then_Place} have explored similar topics. Yet a comprehensive end-to-end solution that seamlessly integrates visual content with natural language constraints is still required. Our methodology, leveraging large multi-modal models, is inherently equipped to bridge this gap.
\begin{table*}[t]
    \centering
    \setlength{\tabcolsep}{4pt}
    \caption{Results comparison on the QB-Poster (up) and UC-Poster dataset (down). In both datasets, PosterLLaVa wins the overall comparison, surpassing all previous methods by better trade-offs between different metrics.} \label{result_cons_QB_Poster}
    \begin{tabular}{lcccccccccccc}
            \toprule
            \multirow{2}[4]{*}{\makecell{\textbf{Methods}}} &  \multicolumn{2}{c}{\textbf{Similarity}} & \multicolumn{3}{c}{\textbf{Content-aware}} & \multicolumn{6}{c}{\textbf{Geometric}} &  \textbf{Constraint}\\
            \cmidrule(lr){2-3} \cmidrule(lr){4-6} \cmidrule(lr){7-12} \cmidrule(lr){13-13}
            & \makecell{Image\\FID$\Downarrow$} & \makecell{IoU$\Uparrow$} & \makecell{Uti$\Uparrow$} & \makecell{Occ$\Downarrow$} & \makecell{Rea$\Downarrow$} & \makecell{Val $\Uparrow$} & \makecell{Ove $\Downarrow$} & \makecell{Ali $\Downarrow$} & \makecell{$\text{Und}_{l}\Uparrow$} & \makecell{$\text{Und}_{s}\Uparrow$} & \makecell{VB$\Downarrow$} & \makecell{Vio$\Downarrow$}\\ 
            \midrule
            & \multicolumn{11}{c}{\textbf{QB-Poster dataset}} \\
            DS-GAN \cite{PosterLayout}  & 85.19 & 0.0558 & 0.5048 & 0.4146 & 0.1995 & 1.0000 & 0.1541 & 0.0034 & 0.3094 & 0.1627 & 0.0287 & - \\
            CGL-GAN \cite{CGL_GAN} & 67.10 & 0.0373 & 0.2908 & 0.3904 & 0.1800 & 0.9959 & 0.1375& 
 0.0040 & 0.3726 & 0.0600 & \textbf{0.0956} & - \\
            ICVT \cite{ICVT} & 97.59 & 0.0231 & 0.1121 & 0.3629 & 0.1442 & 0.9599 & 0.4666 & 0.0018 & 0.4673 & 0.3617 & 0.2903 & - \\
            LayoutDM \cite{LayoutDM} & 159.3 & 0.0144 & 0.2218 & 0.4096 & 0.1850 & 0.9980 & 0.2240 & 0.0003 & 0.4736 & 0.3618 & 0.1223 & - \\
            LayoutPrompter \cite{LayoutPrompter} & 96.86 & 0.0195 & 0.2467 & 0.4504 & 0.1956 & 0.9509 & 0.0233 & 0.0004 & 0.2686 & 0.1501 & 0.2784 & - \\
            \rowcolor{gray!25}
            \textbf{PosterLLaVa(Ours)} & \textbf{35.97} & \textbf{0.1996} & 0.2656 & \textbf{0.3377} & \textbf{0.1659} & 0.9949 & 0.0117 & \textbf{4.75e-5} & \textbf{0.9418} & \textbf{0.9141} & 0.1221 & - \\
            \midrule
            & \multicolumn{11}{c}{\textbf{UC-Poster dataset}} \\
            LayoutPrompter \cite{LayoutPrompter} & 20.29& 0.0961& 0.2024& 0.2846& 0.1038& 0.8512& 0.0014& 0.0018& 0.3916& 0.2906& 0.0781& 0.4130 \\
            \rowcolor{gray!25}
            \textbf{PosterLLaVa(Ours)} & \textbf{3.823}& \textbf{0.1996} & 0.1751& \textbf{0.0924}& \textbf{0.1000}& \textbf{0.9432}& \textbf{0.0014}& \textbf{0.0003}& \textbf{0.9962}& \textbf{0.9944}& \textbf{0.0662} & \textbf{0.1156} \\
            \bottomrule
        \end{tabular}
\end{table*}
\begin{figure}[t]
    \centering
    \includegraphics[width=0.95\linewidth]{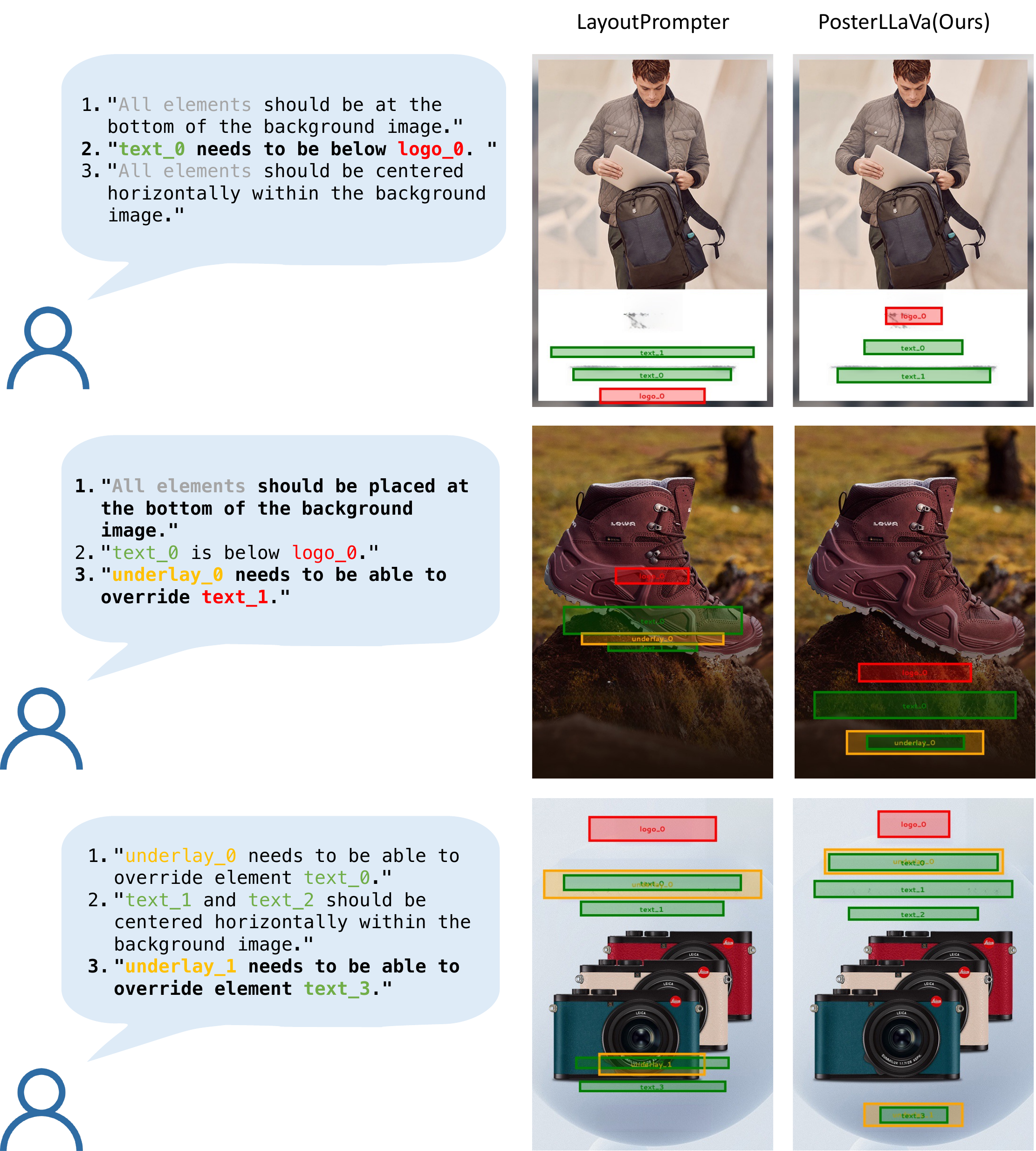}
    \caption{Qualitative results on the User-constrained Poster dataset. The user requirement texts are shown on the left side, and the bold requirement means it was violated by either method.}
    \vspace{-20pt}
\end{figure}

To this end, we propose a new dataset to validate the constrained generation ability of our approach. Firstly, we ask human annotators to write 3 user constraints according to the original poster layout in the CGL \cite{CGL_GAN} validation set (6,006 samples), which are later used as test samples in this experiment. Then, with these high-quality human-annotated constraints serving as in-context learning examples, we utilize ChatGPT to generate constraints automatically. This approach enables us to expand our constraint dataset to include the entire training corpus of the CGL dataset and the PosterLayout dataset, thereby assembling an enormous training dataset of 64,520 samples to mirror the diverse demands of real-world graphic design tasks. The synthesized user-constrained poster dataset is called \textbf{UC-Poster} for short.

\textbf{A New Real-world Poster Dataset} A notable limitation of existing content-aware datasets is their oversimplification. Typically, these datasets feature layouts with no more than 15 design elements, categorized into fewer than 5 types. Such simplicity falls short of conveying sufficient semantic information and mirroring the complexity of designs employed in real-world graphic designs.

To better align with the demands of real-life scenarios, we collect a new dataset named \textbf{QB-Poster} with a much more complicated style. 
As shown in Tab. \ref{tab_dataset},
the elements per poster and geometric complexity of QB-Poster surpass other datasets significantly. This includes 5,188 poster-layout pairs, with 4,675 for training and 513 for testing. The dataset categorizes design elements into 10 categories: title, subtitle, item logo, item, item title, object, text background, decoration, frame, and text. These fine-grained class labels reveal the design pattern of elements and provide the algorithm with additional semantic information. Text elements are organized using a hierarchical classification to indicate their levels of importance. Meanwhile, visual elements are categorized as decoration, text background, object, and frame, which respectively identify decorative icons, underlays, semantically significant objects within background images, and the canvas area.

\textbf{Baseline and Evaluation Metrics} 
We include a wide range of previous content-aware layout generation approaches to ensure a comparison on the proposed QB poster benchmark, including GAN-based~\cite{PosterLayout, CGL_GAN}, VAE-based~\cite{ICVT}, diffusion-based~\cite{LayoutDM}, and LLM-based (empowered by in-context learning)~\cite{LayoutPrompter}. To reproduce LayoutPrompter\cite{LayoutPrompter}, we use the \textit{gpt-3.5-turbo-instruct} instead of the \textit{text-davinci-003} (advised by the original paper) since the latter has been abandoned by OpenAI. For comparison on the UC-Poster dataset, we only include LayoutPrompter since other methods cannot support textual user requirements as input. We extend the original LayoutPrompter by concatenating the extracted saliency bounding box and the constraint texts as inputs to take both background semantics and user requirements into consideration. 

For evaluation metrics, we integrate the evaluation metrics used in public datasets. Namely, we adopt \textit{Image FID} and the bounding box \textit{IoU} from LayoutDETR\cite{LayoutDETR} to measure similarity between generated layouts and the ground-truth layouts; \textit{utility}, \textit{occlusion}, and \textit{readability} from PosterLayout\cite{PosterLayout} to measure content-aware quality; \textit{validity}, \textit{overlap}, \textit{alignment}, \textit{underlay (loose)}, \textit{underlay (strict)}, and \textit{VB (visual balance)} from PosterLayout and HPCVTG\cite{HPCVTG} to measure geometric quality. Most importantly, to measure whether the model follows the input constraints, we sample a subset (50 layouts) of the test set and ask human annotators to verify the average violation ratio of the constraint, marked as \textit{violation}. The overall result is shown in Tab~\ref{result_cons_QB_Poster}.


\textbf{Result Comparison} 
Our proposed PosterLLaVa generally outperforms others in comparisons on the QB-Poster and UC-Poster datasets. In QB-Poster, though some traditional methods win in some particular metrics, our PosterLLaVa achieves the best overall trade-offs. In UC-Poster, our method surpasses LayoutPrompter by a clear margin.

The overall comparison shows that our method excels especially in similarity measurements, the utility/occlusion trade-off, and underlay measurements. 
Therefore, PosterLLaVa tends to place text boxes in the corresponding areas with background color and dialogue bubbles, roughly with an accuracy of 90\%. 
The vision encoder enables this capability and is crucial for developing the poster generation system, PosterGen, which will be discussed in the subsequent section.


\begin{figure*}[t]
    \centering
    \includegraphics[width=\textwidth]{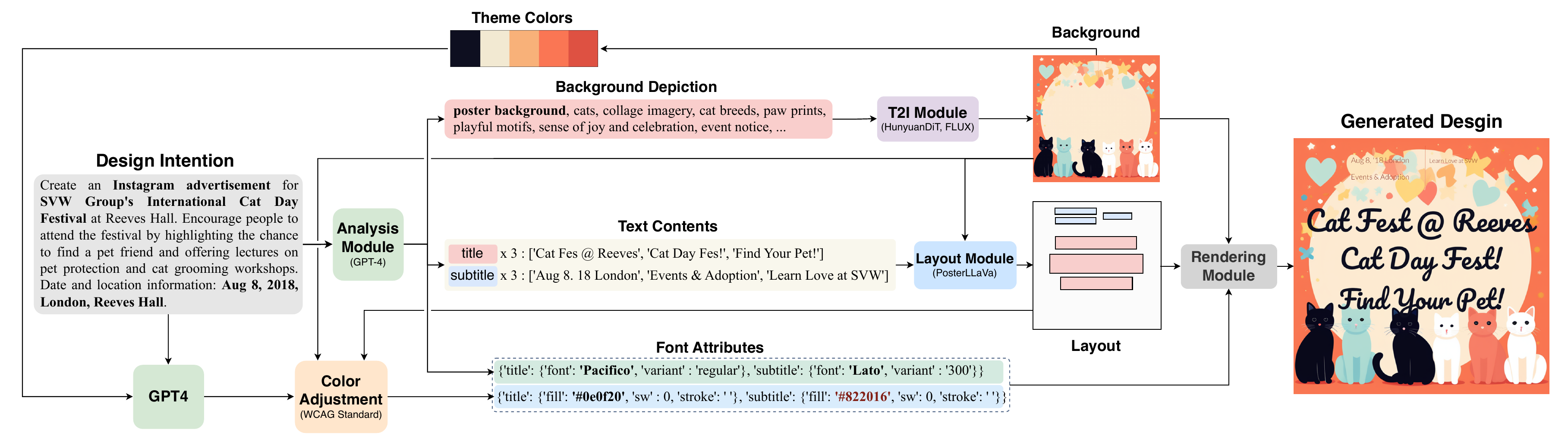}
    \caption{The text-to-poster generation system PosterGen, in which graphic design generation can be decomposed into a) intention analysis, b) text-to-image background generation, c) content-aware layout generation, and d) text attributes generation (font, color, etc.).  As illustrated, PosterGen can correctly interpret the user's design intention to create high-quality background images and display key information in an attention-grabbing location.} \label{fig_pg_framework}
    \vspace{-10pt}
\end{figure*}

\section{Ablation Studies}
We design several ablation experiments to verify the necessity of our proposed method on the following dimensions. We assume that 1. Considering the small scale of the existing content-aware dataset ($<$100,000 samples), the generation performance of the model is positively correlated to the number of training samples and model size; 2. the multi-modal information used should contribute to the generated layout quality. The ad banner dataset is selected for ablation because it is the most lightweight but still contains sufficient multi-modal information, and the metrics used are stable (in contrast, the reliability of utility and occlusion scores highly depends on the quality of saliency detection).

\textbf{Result} The result shown in Tab.~\ref{result_abalation} demonstrates the assumption proposed above. For extra training data, we apply the whole training set of CGL, PosterLayout, and ad banner datasets (78,194 samples in total) for fine-tuning, which improves all geometric measurements. Surprisingly, it also improves the similarity metrics except for Layout FID, which reveals the generality in content-aware generation datasets. Furthermore, the similarity measurement continues to increase by upgrading the pre-trained LLaVa model from 7B to 13B. For multi-modal information, we reduce the visual input (i.e., background image) and textual input (i.e., text element content), respectively, and both of these degrade the overall performance (with a slight improvement in overlap metric, probably because the reduction of information has lower the learning difficulty). These results together demonstrate the effectiveness of utilizing \textbf{multi-modal} \textbf{large models} in content-aware layout generation tasks, and with whose enormous learning capacity, the corresponding demand for \textbf{more high-quality layout data}.

\begin{table}[t]
    \footnotesize
    \centering
    \caption{Ablation Studies conducted on ad banner dataset \cite{LayoutDETR}. Results demonstrate the necessity of applying large models, large datasets, and multi-modal information in content-aware layout generation. } \label{result_abalation}
    \setlength\tabcolsep{2pt}
    \begin{tabular}{lcccccc} 
        \toprule
        \multirow{2}[4]{*}{\makecell{\textbf{Methods}}} & \multicolumn{4}{c}{\textbf{Similarity}} & \multicolumn{2}{c}{\textbf{Geometric}} \\
        \cmidrule(lr){2-5} \cmidrule(lr){6-7}
        & \makecell{Layout\\FID$\Downarrow$} & \makecell{Image\\FID$\Downarrow$} & \makecell{IoU\\$\Uparrow$} & \makecell{DocSim\\$\Uparrow$} & \makecell{Overlap\\$\Downarrow$} & \makecell{Misalign\\$(\times10^{-2})\Downarrow$} \\
        \midrule
        \rowcolor{gray!25}
        \textbf{PosterLLaVa(Ours)} & \textbf{2.37}& 24.87& 0.242& 0.158& 0.029& \underline{1.161}\\
        \midrule
         + extra training data & 3.91& \underline {24.40}& \underline {0.251}& \underline {0.160}& 0.027 &\textbf{0.949}\\
         + 7B$\rightarrow$13B LLM & \underline{2.78}& \textbf{23.86} & \textbf{0.262} & \textbf{0.156} & 0.026 & 1.676  \\
        \midrule
        - textual info & 2.98 & 25.14 & 0.225 & 0.115 & \underline{0.021}& 1.522\\
        - visual info & 8.27 & 40.59 & 0.092 & 0.115 & \textbf{0.020} & 2.193\\
        \bottomrule
    \end{tabular}
    \vspace{-20pt}
\end{table}

\section{PosterGen: An Automated Text-to-Editable Poster System with Multi-language Support}

\begin{figure*}[t]
    \centering
    \includegraphics[width=\textwidth]{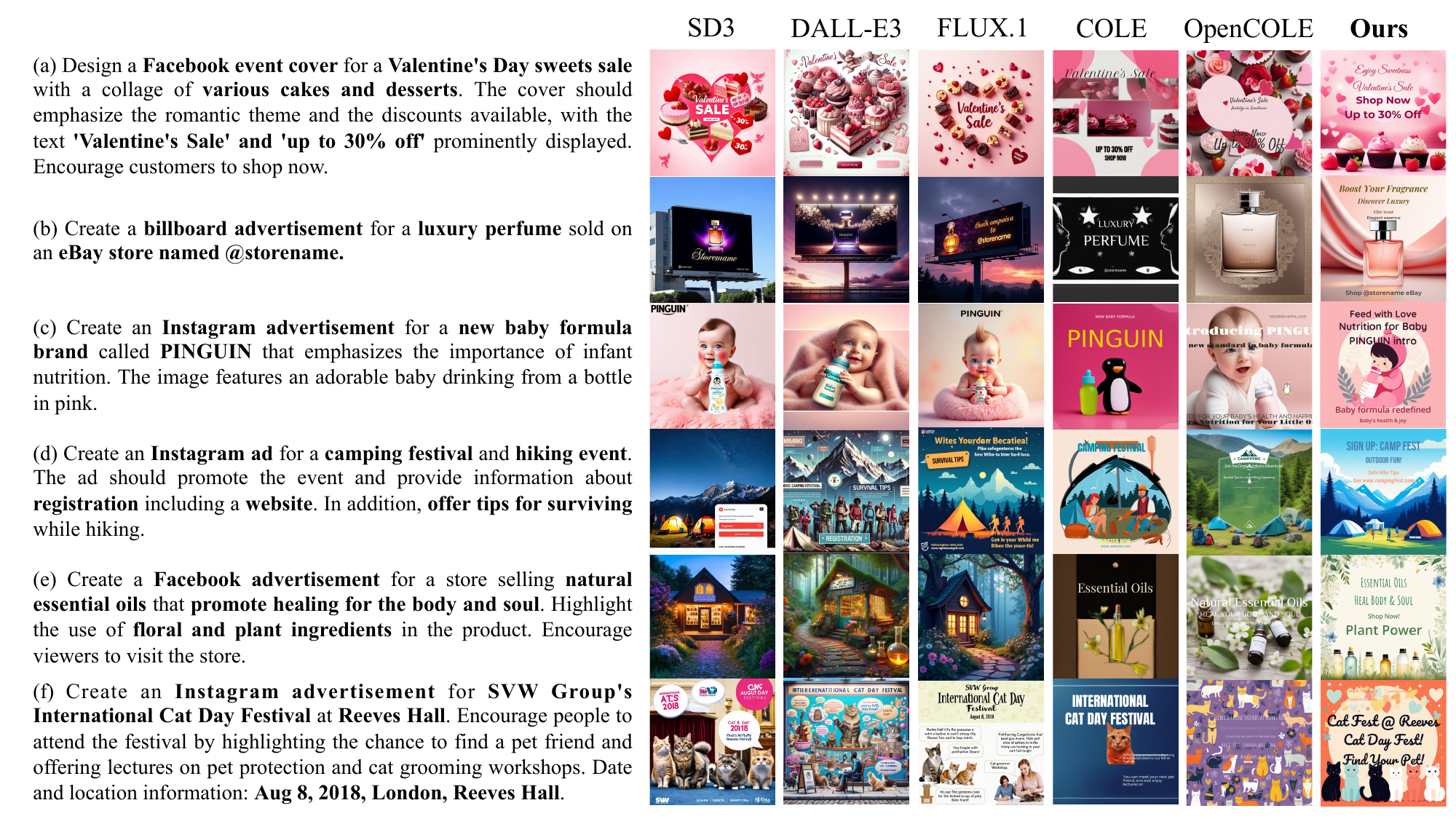}
    \caption{The qualitative comparison on DESIGNERINTENSION~\cite{cole} benchmark with recently proposed poster generation method COLE \cite{cole} and OpenCOLE \cite{opencole}, and text-to-image models\cite{sdxl, dalle3, flux} (GPT-4 based prompt augmentation are adopted as advised by \cite{cole}). Our method has better editability than vanilla T2I schemes and outperforms competitors with better background quality, text readability, and fewer training resources.} \label{fig_di_bench}
    \vspace{-10pt}
\end{figure*}
\begin{figure*}[t]
    \centering
    \includegraphics[width=0.85\textwidth]{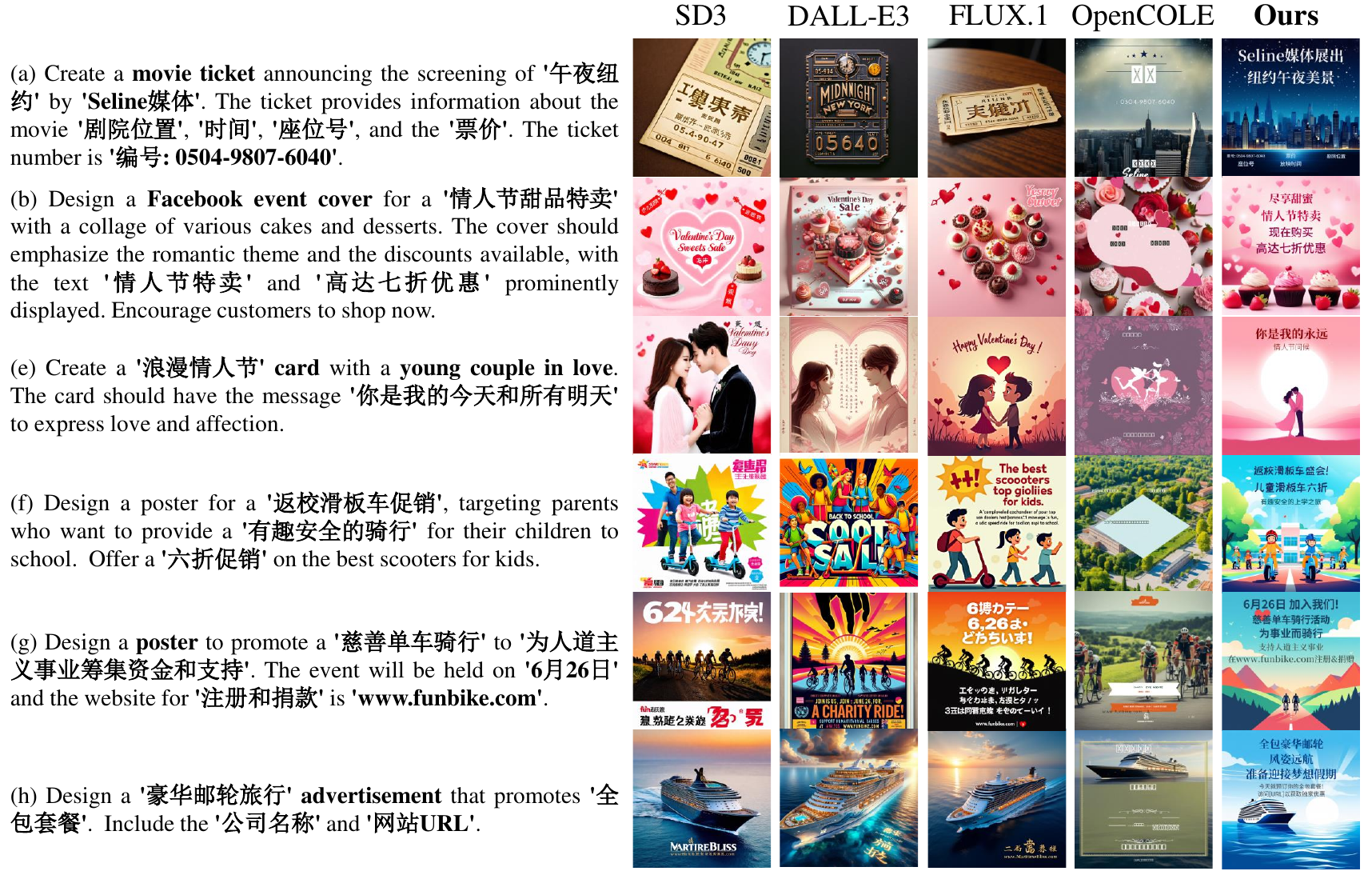}
    \caption{The qualitative comparison of Chinese text illustration, showcasing the multi-lingual adaptability of the proposed method. COLE \cite{cole} is excluded due to its public inaccessibility. Latest T2I methods \cite{sd3, dalle3, flux} still cannot produce meaningful Chinese characters; OpenCOLE \cite{opencole} fails to render since the fonts they included cannot support Chinese characters; while our method stably generates high-quality samples.} \label{fig_zh_bench}
    \vspace{-10pt}
\end{figure*}

In Section~\ref{sec_method}, we propose PosterLLaVa to tackle the challenge of layout generation given the background image and element categories and validate its effectiveness on a wide range of datasets. 
However, in real-world applications, the bounding-box-based representation is rarely the final goal of automated art design, because layout generation is only a part of automotive art design. In real-world scenarios, the business pipeline often begins with vague design requirements (such as user intent, product descriptions, news, or event announcements) and ends with a complete poster design, which needs the teamwork of product managers, graphic designers, front-end programmers, etc. and costing a lot of time and human effort. Despite this pressing demand, existing works \cite{CGL_GAN, PosterLayout, LayoutDiffusion, LayoutDM}, focusing on polishing the layout generation algorithm, provide limited insight into the downstream transition from layout to complete poster design. There is a severe lack of accessible open-source tools to facilitate this process and save human effort in the process of graphic designing. 

Noticing this fact, we further developed a poster generation system (named PosterGen) based on PosterLLaVa to facilitate automated text-to-poster design. The overall framework of PosterGen is shown in Fig.~\ref{fig_pg_framework}. The system can be generally divided into 3 stages: 1) intention analysis: our system utilizes an analysis module (i.e., GPT-3.5  in our implementation) to analyze the background information and the intention given by the user. This step will refine these intentions into three groups of more detailed information: prompts for background generation; copy-writings and categories, namely the text elements to be presented on the design; and font suggestions, including font family and variant. 2) Background generation: subsequently, a pre-trained text-to-image diffusion model is adopted to generate a high-quality background using the refined prompts. 3) Layout generation: with the background image, copywriting, and corresponding categories, we adopt our PosterLLaVa to produce reasonable layout suggestions. 4) Font color selection: we extract the theme color from the generated background with clustering and call the GPT-3.5 again for harmonious text color suggestions. These initial colors are then filtered and fine-tuned according to the lightness distribution of their corresponding background patches (see Appendix for details of WCAG standard-based color adjusting scheme) to improve readability. Our observation finds that font types can be directly inferred from the intention. Differently, these color selection steps are separated independently because they are related to multi-modal information and thus cannot be paralleled or integrated into step one like font selection. Assuming the copywriting doesn't involve line breaks, the optimal font size can be inferred using the bonding box (containing size and location), text content, and font type. Finally, the rendering module combines all the above information to create a text-overlaid poster image. 

\subsection{Key Observations} Our system is built upon two key empirical observations about the zero-shot inference capability of large models: 1) some diffusion models (e.g., HunyuanDiT \cite{hunyuandit}, FLUX \cite{flux}) can generate background images with placeholders to facilitate the text illustration step later, simply by inserting special tokens like "poster background", and "blank-leaving" in prompts. 2) Some LLMs (e.g. GPT \cite{GPT4}, LLaMa3 \cite{LLaMa3}) have preliminary knowledge of public-available fonts. This includes distinguishing whether a given font belongs to the serif or sans-serif category, handwritten or printed style, and assessing the suitability of these categories for the current theme, which is sufficient to make reasonable font choices considering the given design intention and audiences' preference. These two observations allow us to build our PosterGen system with PosterLLaVa as the only fine-tuned model, minimizing the need for expensive training procedures. In contrast, recently proposed approaches (COLE \cite{cole} and OpenCOLE \cite{opencole} recommend fine-tuning both the diffusion model and the LLM to solve background generation and typography selection, requiring numerous computational and data resources. In the following experiment (see Fig. \ref{fig_radar_chart}), we are surprised to observe that our zero-shot method even surpasses these methods by a clear margin concerning the three dimensions of \textit{graphics \& images}, \textit{design layout}, and \textit{topography \& color}. This demonstrates that the quality of existing graphic design datasets is still far from expectations, and over-relying on them can lead to a degradation in models' generation quality.

\subsection{Qualitative Evaluation}
\textbf{Evaluation with DESIGNERINTENTION Samples} The corresponding qualitative comparisons are shown in Fig.\ref{fig_di_bench}, in which we select 6 intentions (a-f) and show the design results of 6 methods. Generally speaking, T2I methods have better image quality, but their drawbacks are also clear. As suggested by COLE\cite{cole}, for the T2I methods, we apply GPT-4 augmentation to enrich and expand the original design intention into detailed prompts. This method enriches the painting details. However, this method sometimes results in pure images without text like in cases b, c, and e; or, strange image compositions and unnecessary scenes, like in cases b and e. Remarkably, DALLE-3 often generates over-complicated design layouts, like in case a, d, and f, distracting the users' attention. Different from SD-XL\cite{sdxl}, the SOTA T2I methods\cite{sd3, dalle3, flux}, can produce images with clear English titles now, but the small print content is still unrecognizable (like DALLE-3 in case f) or inundated with non-sense typos (like SD3 and FLUX in case d and f). In contrast, the text-to-editable-graphic-design methods are distinctively more readable. The text color selection of COLE and OpenCOLE is stiffness, with black and white appearing in a dominantly large possibility. OpenCOLE chooses extremely thin font types and small font sizes, and meanwhile, the text is often placed out of the placeholder (like in case a and d), further reducing its readability. Our method generally outperforms all the baselines with high image quality and style consistency, reasonable and readable font type/color selection, and attractive layout design with all text elements placed accordingly - benefited by our proposed PosterLLaVa model trained on numerous high-quality layout data.

\textbf{Evaluation of Chinese Samples} One distinct advantage of our method is its multilingual ability, which is a benefit of exploring the zero-shot font selection ability of GPT. We show some qualitative examples in Fig~\ref{fig_zh_bench} for comparison by replacing the copywriting (but NOT the whole intention) with Chinese translations to create posters illustrated with Chinese characters. COLE~\cite{cole} is excluded because of the inaccessibility of source code to reproduce their method. OpenCOLE~\cite{opencole} includes 753 types of fonts, but all with Latin character sets, causing the garbled characters when trying to render Chinese. Though supporting English characters to some extent, the T2I method fails to generate recognizable and meaningful Chinese characters. This is due to the difference in the total number of different languages and has been discussed in previous works~\cite{GlyphDraw, AnyText}. Another interesting phenomenon is that due to data distribution, these T2I methods sometimes cannot distinguish whether the user wants Chinese glyphs or Chinese styles, like SD3 and FLUX in case a, as well as SD3 and DALLE-3 in case c - in which we are trying to illustrate themes of western concepts but with Chinese characters. Our method undoubtedly wins in the multi-lingual comparison by producing high-quality Chinese posters aligned with various kinds of intentions. For instance, in case f, it chooses to use a cute font to align with the theme of children's product promotion.

\subsection{Quantitative Evaluation}

\begin{figure*}[t]
    \centering
    \includegraphics[width=\textwidth]{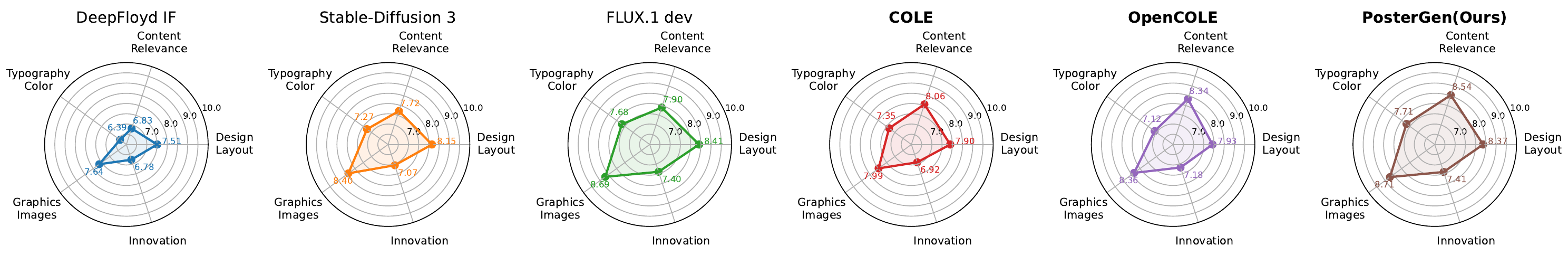}
    \caption{The quantitative comparison (with GPT-4o evaluated scores) between SOTA text-to-image commercial models~\cite{sd3, flux, dalle3}, SOTA graphic designer generation approaches~\cite{cole, opencole} and our method on DESIGNERINTENSION~\cite{cole} benchmark. The proposed PosterGen generally outperforms others, with the simplest pipeline and lowest training efforts.} \label{fig_radar_chart}
    \vspace{-10pt}
\end{figure*}
\begin{figure*}[t]
    \centering
    \includegraphics[width=0.98\textwidth]{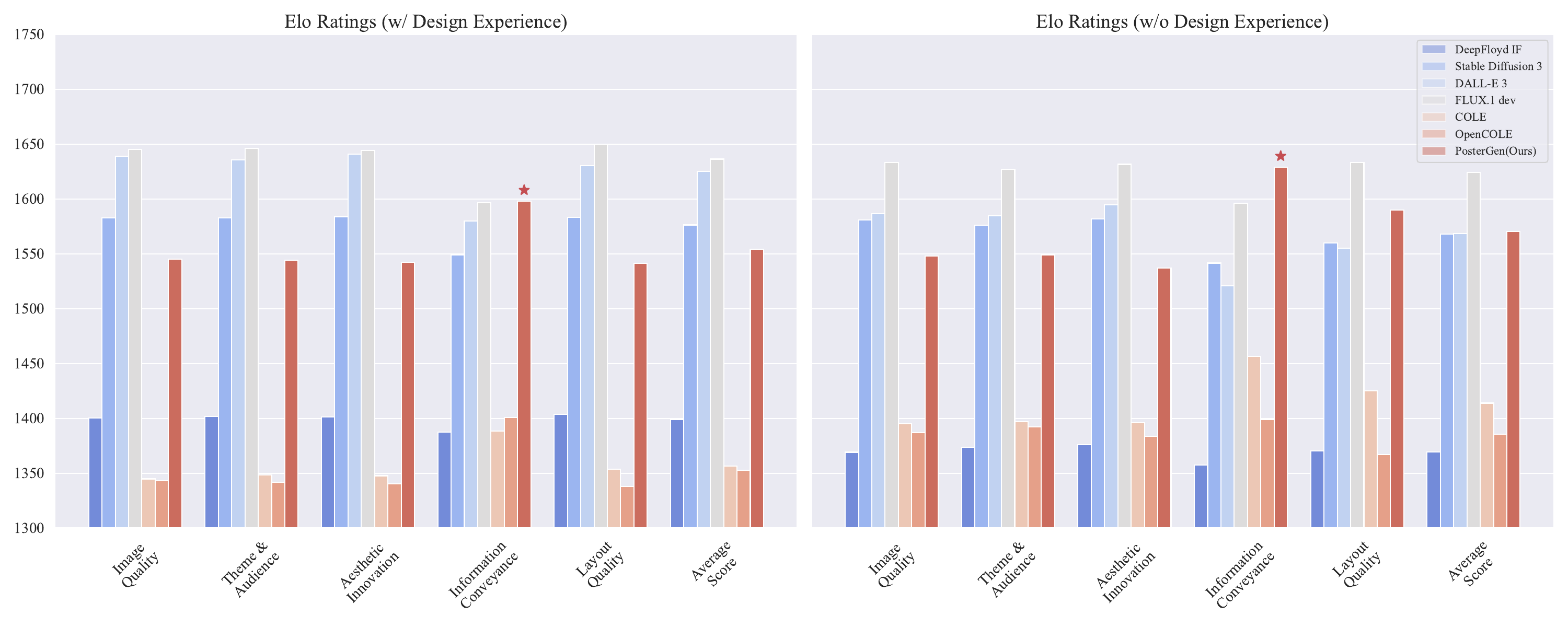}
    \vspace{-10pt}
    \caption{The quantitative comparison on DESIGNERINTENSION~\cite{cole} benchmark through human-annotated user studies. We sample different intentions and methods to create pair-wise comparison and compute the ultimate score with Elo\cite{Elo} rating mechanism.} \label{fig_bar_chart}
    \vspace{-10pt}
\end{figure*}

\textbf{GPT-4o Evaluations} The evaluation of graphic design is nontrivial, due to its subjectivity and non-uniqueness. COLE \cite{cole} provides the DESIGNERINTENTION benchmark, which contains 500 textual design intentions from different areas (e.g., ad, poster, social media). OpenCOLE \cite{opencole} opensource the evaluation code using GPT-4 and suggests evaluating on a 200-sample subset of the benchmark. We generally follow the evaluation pipeline advised by OpenCOLE: render graphic 
 designs into images and use GPT-4o \cite{GPT4} to evaluate the quality of the graphic designs to improve efficiency and save expenses. The comparison includes five dimensions: a) \textit{design and layout}, b) \textit{content relevance and effectiveness}, c) \textit{typography and color scheme}, d) \textit{graphic and images}, and e) \textit{innovation}. COLE and OpenCOLE are included as SOTA graphic design generation methods for comparison, T2I methods with enhanced text rendering ability \cite{deepfloyd, sd3, dalle3, flux} are also taken into consideration, even though they cannot produce veritable graphic designs, but generate non-editable picture prototypes instead. According to the experiment result shown in Fig~\ref{fig_radar_chart}, PosterGen wins the comparison of all five dimensions among text-editable methods (shown in bold font), especially in graphics and images, layout, and typography. Our method achieved a notable high score (that is, 8.54) in terms of content relevance, indicating its ability to analyze intent and user preferences. As a pure T2I model, FLUX performs surprisingly well considering layout and topography, which are, unfortunately, encoded in pixel images implicitly and cannot be easily extracted or edited. All included methods have room for improvement in terms of innovation. Noted that we ignore DALLE3 in Fig/\ref{fig_radar_chart}  it's hard to use GPT-4o to evaluate images generated by DALLE3\cite{dalle3}, which seems to have some inner specific safety checker and fails to decode with a possibility of over 80\%.

\textbf{User Studies} We extend the quantitative evaluation by conducting human-based user studies. We hire two groups of people on the crowd-sourcing platform, including 15 annotators. They are divided into two groups, 5 are with graphic design experience and 10 are not. Besides, different from giving absolute scores ranging from 1 to 10 as GPT did, we randomly sampled image pairs and asked human subjects to compare and choose the winner. We believe the pair-wise comparison is more intuitive for humans to perform. Finally, we compute the Elo\cite{Elo} score for each model regarding 5 dimensions, related details are discussed in the appendix. As illustrated in Fig. \ref{fig_bar_chart}, we can have the following observations: 1. Due to numerous high-quality training data, the overall quality of the latest T2I method is conspicuously superior to the graphic design generation method, among which FLUX is the most remarkable one, winning in most comparisons. 2. The average quality of our method is better than DeepFloyd IF\cite{deepfloyd}, and approximately compatible with SD3\cite{sd3} and DALLE-3\cite{dalle3}, which constitutes the best average-quality graphic design generation method. Moreover, its information communication is extremely prominent, even outperforming FLUX\cite{flux} in both the design and non-design groups; its layout quality surpasses SD3 and DALLE3 in the non-design group evaluation, which reflects the effectiveness of our proposed layout generation model and text attributes polishing algorithm. 3. Regardless of the dimension, the design group has a greater preference for the T2I method than the non-design group, which indicates the necessity of collecting more high-quality graphic design data and enhancing the generation pipeline. Generally speaking, this result is in accord with the previous GPT evaluation (more analysis are included in the appendix).

\section{Conclusion}
Content-aware layout generation is a highly multi-modal problem. Utilizing the latest multi-modal large model instruction fine-tuning techniques, we propose a method named PoserLLaVa that represents multi-modal layout information as tokens, which are then processed by a Large Language Model (LLM). The proposed method achieves SOTA performance across multiple content-aware layout generation datasets. Additionally, by surveying existing content-aware layout generation datasets, we identify significant shortcomings in the current public datasets, namely the lack of user-constrained data and complicated data, both of which are crucial in real-world applications. We further collect two new datasets to bridge this gap, the user-constrained poster dataset and the QB-Poster, based on which we verify the extended ability of our method. In summary, to achieve large-scale automated production, high-quality multi-modal layout data and a unified learning approach are still under demand, for which our method paves the way.

\ifCLASSOPTIONcaptionsoff
  \newpage
\fi



%

\bibliographystyle{IEEEtran}
\bibliography{reference}

\end{document}


%
%
%
%
\title{PosterLLaVa: Constructing a Unified Multi-modal Layout Generator with LLM}

\IEEEdisplaynontitleabstractindextext

\maketitle
%
\IEEEpeerreviewmaketitle

\appendix

\subsection{Details for Quantitative Evaluation with GPT-4o} \label{gpt_4o_eval}

\textbf{Evaluation Prompt} We generally follow the prompt suggested by COLE\cite{cole} in absolute evaluation and assign every graphic design with scores of 1-10 regarding 5 dimensions.

\begin{figure*}[t]
    \centering
    \includegraphics[width=\linewidth]{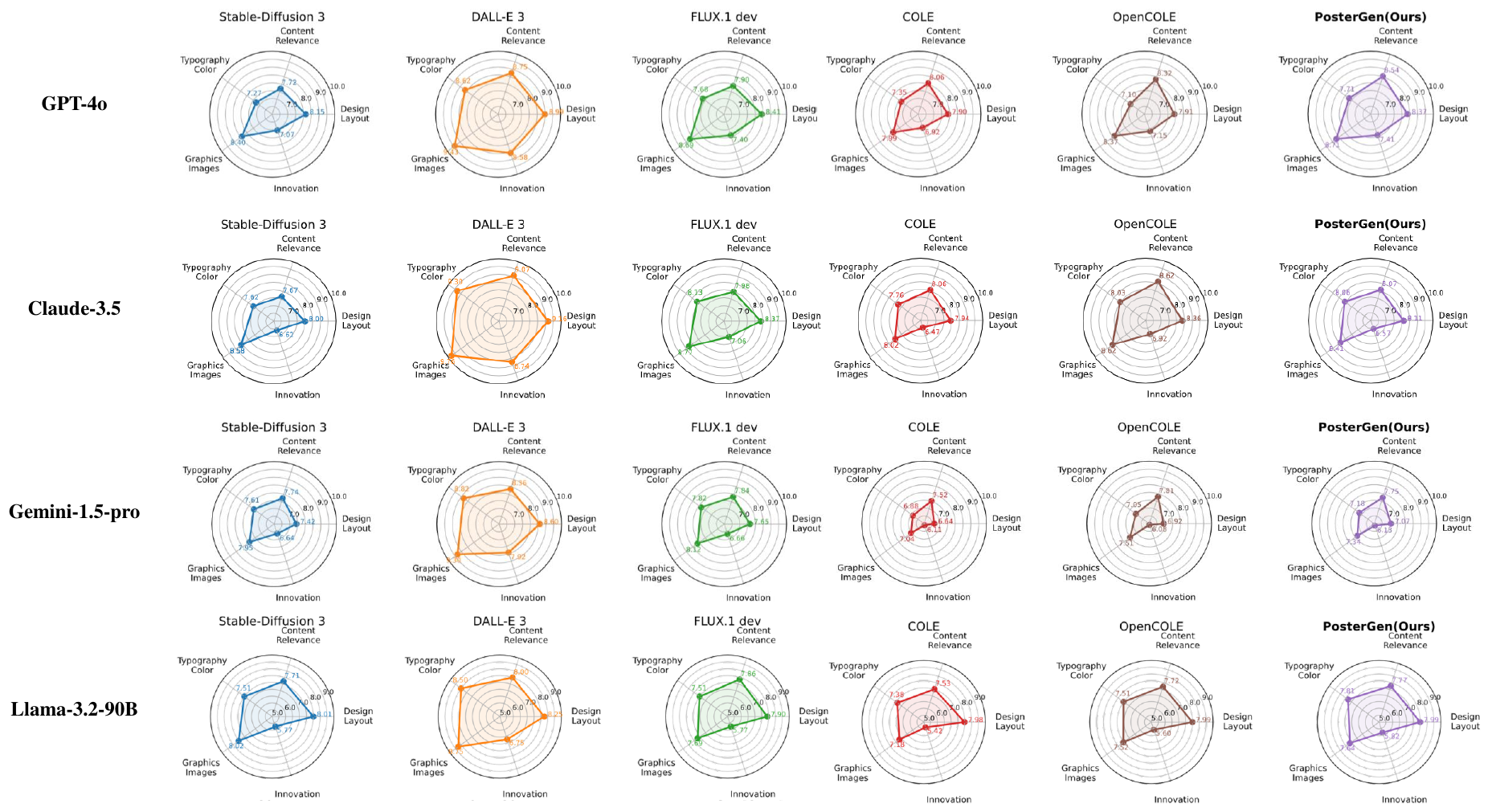}
    \caption{Qualitative results with more VLLMs~\cite{gpt4o, Gemini1d5, claude35sonnet, LLaMa3}.}
    \label{more_vllms_radar}
\end{figure*}

\textbf{More VLLMs Evaluation} Apart from GPT-4o\cite{gpt4o}, there are recently other Large Language model alternatives with similar visual ability \cite{Gemini1d5, claude35sonnet, LLaMa3}. Another reason for considering this is that it's possible that OpenAI has already trained GPT-4o with DALLE3\cite{dalle3} generated images or used GPT-4o as the T2I interpreter before the diffusion model of DALLE3; both of these situations would result in unfair comparisons. Thus, we choose Gemini 1.5 pro, Claude-3.5-sonnet-v2, and LLaMa-3.2-90B for a border comparison. The results are shown in Fig.~\ref{more_vllms_radar}.

\begin{table*}[t]
    \centering
    \begin{tabular}{p{0.95\textwidth}}
    \toprule
    \textsc{\textbf{User:}} \\
    \textsf{\text{$<$image$>$}}\\
    \\
    You are an autonomous AI Assistant who aids designers by providing insightful, objective, and constructive critiques of graphic design projects. Your goals are: Deliver comprehensive and unbiased evaluations of graphic designs based on established design principles and industry standards. Identify potential areas for improvement and suggest actionable feedback to enhance the overall aesthetic and effectiveness of the designs. Maintain a consistent and high standard of critique. Utilize coordinate information for data description relative to the upper left corner of the image, with the upper left corner serving as the origin, the right as the positive direction, and the downward as the positive direction. Please abide by the following rules: Strive to score as objectively as possible. Grade seriously. A flawless design can earn 10 points, a mediocre design can only earn 7 points, a design with obvious shortcomings can only earn 4 points, and a very poor design can only earn 1-2 points. Keep your reasoning concise when rating, and describe it as briefly as possible. If the output is too long, it will be truncated.\\
    \\
    Only respond in JSON format, no other information. Example of output for a perfect design: \{"design\_and\_layout": 10, "content\_relevance\_and\_effectiveness": 10, "typography\_and\_color\_scheme": 10, "graphics\_and\_images": 10, "innovation\_and\_originality": 1\}' \\
    \\
    Grading criteria\\
    Design and Layout (name: design\_and\_layout, range: 1-10): The graphic design should present a clean, balanced, and consistent layout. The organization of elements should enhance the message, with clear paths for the eye to follow. A score of 10 signifies a layout that maximizes readability and visual appeal, while a 1 indicates a cluttered, confusing layout with no clear hierarchy or flow. 
    Content Relevance and Effectiveness (name: content\_relevance\_and\_effectiveness, range: 1-10): The content should be not only relevant to its purpose but also engaging for the intended audience, effectively communicating the intended message. A score of 10 means the content resonates with the target audience, aligns with the design’s purpose, and enhances the overall message. A score of 1 indicates the content is irrelevant or does not connect with the audience.\\
    Typography and Color Scheme (name: typography\_and\_color\_scheme, range: 1-10): Typography and color should work together to enhance readability and harmonize with other design elements. This includes font selection, size, line spacing, color, and placement, as well as the overall color scheme of the design. A score of 10 represents excellent use of typography and color that aligns with the design’s purpose and aesthetic, while a score of 1 indicates poor use of these elements that hinders readability or clashes with the design.\\
    Graphics and Images (name: graphics\_and\_images, range: 1-10): Any graphics or images used should enhance the design rather than distract from it. They should be high quality, relevant, and harmonious with other elements. A score of 10 indicates graphics or images that enhance the overall design and message, while a 1 indicates low-quality, irrelevant, or distracting visuals.\\
    Innovation and Originality (name: innovation\_and\_originality, range: 1-10): The design should display an original, creative approach. It should not just follow trends but also show a unique interpretation of the brief. A score of 10 indicates a highly creative and innovative design that stands out in its originality, while a score of 1 indicates a lack of creativity or a generic approach. \\
    \\
    \textsc{\textbf{Assistant:}}\\
    \bottomrule
    \end{tabular}
    \vspace{10pt}
    \caption{Prompt template for applying visual instruction tuning on content-aware generation task. The placeholder tokens in bold type are replaced with specific information during training or inference.} \label{gpt_4o_prompt_template}
\end{table*}

\subsection{Details for Quantitative Evaluation with Human Subjects}
\textbf{Evaluation Dimensions} We slightly modified the original 5 comparison metrics in Section~\ref{gpt_4o_eval} to obtain the dimensions we used for user studies. Specifically, we use "Image \& Quality," "Theme \& Audience," "Aesthetic \& Innovation," "Information Delivery," and "Layout Quality.“ 
Considering the purpose of graphic poster design, we especially highlight Information Delivery as a strengthened version for Text Rendering \& Color scheme. This not only requires appropriate and harmonious font rendering but also the effective conveyance and presentation of the key information mentioned in the design intention. 

\begin{figure}[t]
    \begin{subfigure}
        \centering
        \includegraphics[width=\linewidth]{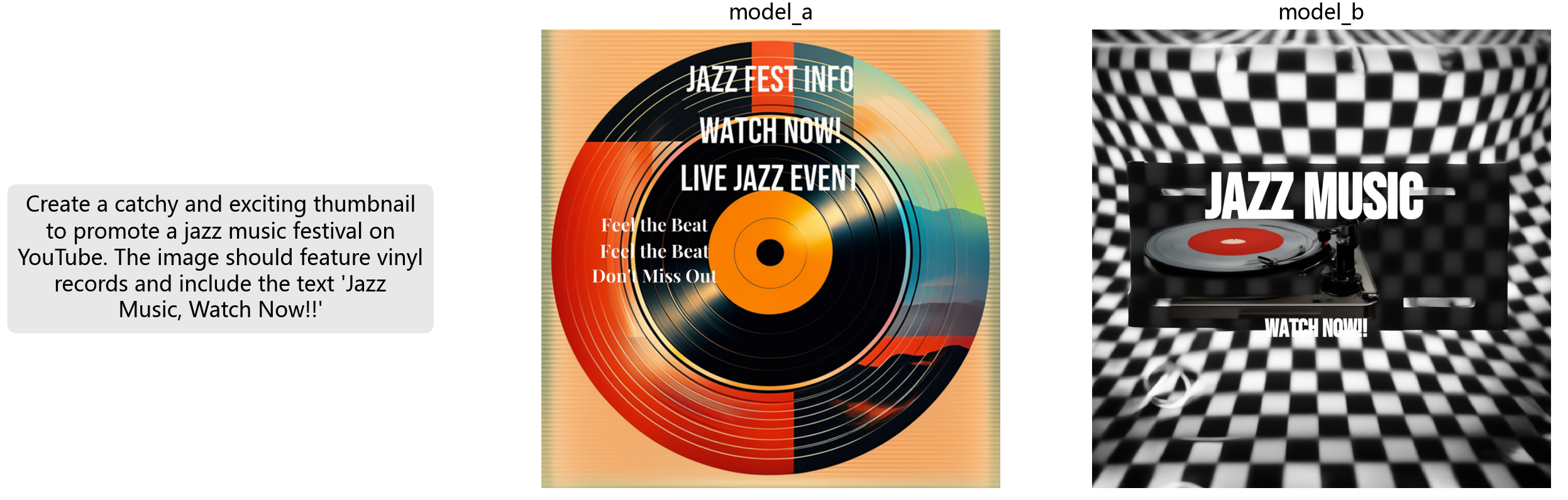}
    \end{subfigure}
    \begin{subfigure}
        \centering
        \includegraphics[width=\linewidth]{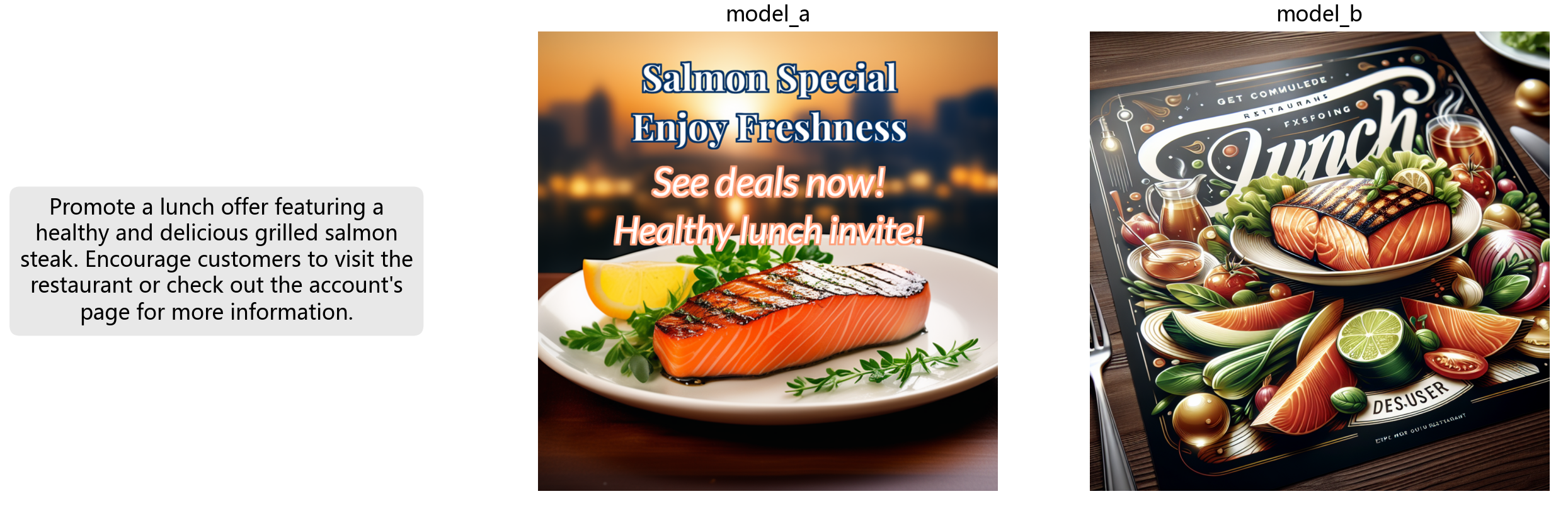}
    \end{subfigure}
    \begin{subfigure}
        \centering
        \includegraphics[width=\linewidth]{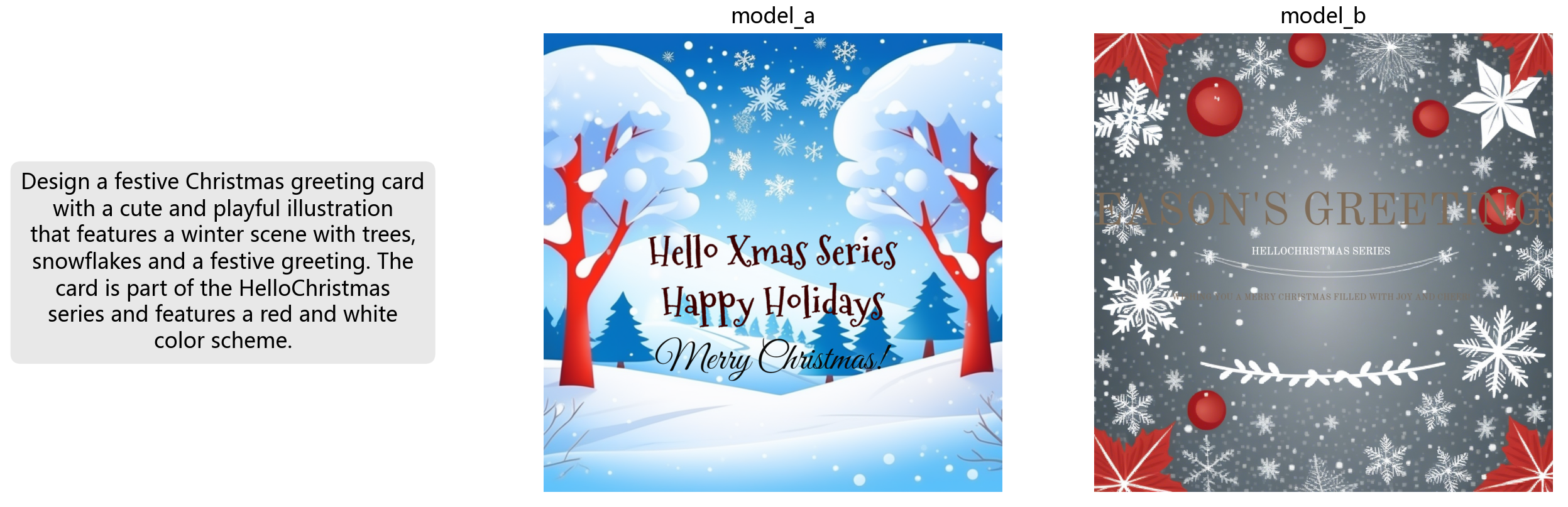}
    \end{subfigure}
    \caption{Pairwise comparison samples, for which intentions and baseline pairs are randomly sampled to ensure a fair evaluation. During evaluation, method names are erased to reduce annotators' prejudice against specific methods.}
    \label{user_study_samples}
    \vspace{-10pt}
\end{figure}

\textbf{Pair-wise Comparison} As discussed in Section~\ref{gpt_4o_eval}, we employ GPT-4o to assign an absolute score to each design according to the given criteria. However, applying this approach directly to user studies presents challenges. Prior research \cite{rossi2001overcoming, chen2009crowdsourceable, shen2016ranking} has highlighted that pairwise comparison is generally superior to absolute scoring in crowd-sourced quality assessments, considering it enhances subjective consistency and reduces evaluator workload. Consequently, we sample the image pairs in the 200-sample subset of DESIGNERINTENTIONS, each time only with 2 baselines for comparison. We include 6 baselines: DeepdFloy IF, Stable Diffusion 3, DALLE-3, FLUX.1 dev, COLE, OpenCOLE, and our methods for a comprehensive evaluation, which results in $\binom{7}{2}$ kinds of pairwise combinations, thus the total image pairs are $200\times\binom{7}{2}=4200$. For the non-design group with 10 people, each is dedicated to 420 pairs of samples; and for the design group with 5 people, each is 840 pairs. Figure~\ref{user_study_samples} presents several evaluation examples where human evaluators are asked to select the preferred model between the two.

\textbf{Elo Rating Scheme} Several methods \cite{elo, glicko, glicko2, trueskill, Bradley_Terry} exist for converting pairwise (or multi-player) comparison results into scores for each method, with the Elo rating system \cite{elo} being the most widely recognized. Originally developed for Chess and Go competitions, the Elo system calculates the expected probability of victory for two players, A and B, using the following formula:
$$
    E_A = \frac{1}{1+10^{(R_B-R_A)/400}}, \quad E_B = \frac{1}{1+10^{(R_A-R_B)/400}}
$$
where $R_A$ and $R_B$ are the ratings of players A and B, respectively. Upon observing a new competition result, the players' ratings are updated as follows:
$$
    R_{A}^{\prime} = R_A + KE_B, \quad R_{B}^{\prime} = R_B - KE_A 
$$ assuming player A wins, and vice versa. $K$ is a positive constant representing the maximum possible rating adjustment per game and controls the magnitude of score changes. In our experiment, we initialize $K=20$.

However, directly applying the Elo rating system to compute final scores can lead to issues, as Elo ratings are sensitive to the sequence of competitions. Since our comparisons are randomly synthesized, the order in which competitions occur should not influence the results. To address this, we propose an iterative sampling scheme for modified Elo ratings computation. Specifically, we initialize Elo scores of all methods with 1500. Then, we randomly sample a batch of comparison results (without replacement) and update the Elo ratings of different methods accordingly. In each subsequent iteration, we multiply the K-value by a factor of $0.95$ so that the magnitude variation of scores is iteratively reduced, and the random sampling is maintained to alleviate the influence of competition order. In our case, Elo scores of different methods converge within an error of $1e-3$ after $177$ iterations. We also found empirically that the sampling order in the early iterations does not significantly affect the final scores either.

\subsection{Details for Color Refinement Algorithm}

Empirically, we found GPT are not enough to select reasonable color purely from design intention. The select of color should, at least follow two rules: color harmony and color contrast. Previous paper 

\begin{table*}[t]
    \centering
    \setlength{\tabcolsep}{4pt}
    \caption{Additional comparison on the QB-Poster dataset. We include different instruction tuning schemes for comparison, but only LLaVa-V1.5 is adopted for results in the main body.} \label{result_addi}
    \begin{tabular}{lccccccccccc}
        \toprule
        \multirow{2}[4]{*}{\makecell{\textbf{Methods}}} &  \multicolumn{2}{c}{\textbf{Similarity}} & \multicolumn{3}{c}{\textbf{Content-ware}} & \multicolumn{6}{c}{\textbf{Geometric}} \\
        \cmidrule(lr){2-3} \cmidrule(lr){4-6} \cmidrule(lr){7-12}
        & \makecell{Image\\FID$\Downarrow$} & \makecell{IoU$\Uparrow$} & \makecell{Uti$\Uparrow$} & \makecell{Occ$\Downarrow$} & \makecell{Rea$\Downarrow$} & \makecell{Val $\Uparrow$} & \makecell{Ove $\Downarrow$} & \makecell{Ali $\Downarrow$} & \makecell{$\text{Und}_{l}\Uparrow$} & \makecell{$\text{Und}_{s}\Uparrow$} & \makecell{VB$\Downarrow$} \\ 
        \midrule
        & \multicolumn{11}{c}{\textbf{QB-Poster dataset}} \\
        DS-GAN \cite{PosterLayout} & 85.19 & 0.0558 & \textbf{0.5048} & 0.4146 & 0.1995 & \textbf{1.0000} & 0.1541 & 0.0034 & 0.3094 & 0.1627 & \textbf{0.0287} \\
        CGL-GAN \cite{CGL_GAN} & 67.10 & 0.0373 & 0.2908 & 0.3904 & 0.1800 & 0.9959 & 0.1375 & 0.0040 & 0.3726 & 0.0600 & 0.0956 \\
        ICVT \cite{ICVT} & 97.59 & 0.0231 & 0.1121 & 0.3629 & 0.1442 & 0.9599 & 0.4666 & 0.0018 & 0.4673 & 0.3617 & 0.2903 \\
        LayoutDM \cite{LayoutDM} & 159.3 & 0.0144 & 0.2218 & 0.4096 & 0.1850 & 0.9980 & 0.2240 & 0.0003 & 0.4736 & 0.3618 & 0.1223 \\
        LayoutPrompter \cite{LayoutPrompter} & 96.86 & 0.0195 & 0.2467 & 0.4504 & 0.1956 & 0.9509 & 0.0233 & 0.0004 & 0.2686 & 0.1501 & 0.2784 \\
        \midrule
        PosterLLaVa (miniGPT-4)& 	114.89& 0.0712& 0.2650& 0.3602& 0.1781& 0.9944& 0.0086& 2.122e-5& 0.9260& 0.9111& 0.1658\\
        PosterLLaVa (mPLUG-owl2)& 59.11& 0.0712& 0.2650& 0.3602& 0.1781& 0.9944& \textbf{0.0086}& \textbf{2.122e-5}& 0.9260& 0.9111& 0.1658\\
        \rowcolor{gray!25}
        \textbf{PosterLLaVa (LLaVa-v1.5)}& \textbf{35.97} & \textbf{0.1996} & 0.2656 & \textbf{0.3377} & \textbf{0.1659} & 0.9949 & 0.0117& 4.75e-5& \textbf{0.9418} & \textbf{0.9141} & 0.1221 \\
        \bottomrule
    \end{tabular}
\end{table*}

\begin{table*}[t]
    \centering
    \setlength{\tabcolsep}{2pt}
    \caption{Training expenditure analysis on the QB-Poster training set (4,675 posters). The experiment empirically demonstrates that introducing LLMs under acceptable computational expenditure is feasible.} \label{efficiency_test}
    \begin{tabular}{lccc}
        \toprule
        \textbf{Methods} & \textbf{Training Device} & \textbf{Training Time (sec)} & \textbf{Training Epochs} \\
        \midrule
        & \multicolumn{3}{c}{\textbf{Previous Methods}} \\
        DS-GAN \cite{PosterLayout} & 16 X NVIDIA A10 (24GB)& 9030&300 \\
        CGL-GAN \cite{CGL_GAN} & 16 X NVIDIA A10 (24GB)& 21667&300 \\
        ICVT \cite{ICVT} & 16 X NVIDIA A10 (24GB)& 12030&300 \\
        LayoutDM \cite{LayoutDM} & 16 X NVIDIA A10 (24GB)& 19740&300 \\
        \midrule
        & \multicolumn{3}{c}{\textbf{LLMs-based Methods}} \\
        PosterLLaVa (miniGPT4)& 8 X NVIDIA A100 (40GB) & 3414 & 20 \\
        PosterLLaVa (mPLUG-owl2)& 16 X NVIDIA A10 (24GB) & 1628 & 2 \\
        \rowcolor{gray!25}
        \textbf{PosterLLaVa (LLaVa-v1.5)}& 8 X NVIDIA A10 (24GB) & 4186 & 2 \\
        \rowcolor{gray!25}
        \textbf{PosterLLaVa (LLaVa-v1.5 LoRA)} & 8 X NVIDIA A10 (24GB) & 2093 & 2 \\
        \bottomrule
    \end{tabular}
\end{table*}

\subsection{More Visual Instruction Tuning Schemes}
We also included other instruction-tuning techniques for comparison. MiniGPT-4~\cite{miniGPT4} is an instruction-tuning method based on Q-Former\cite{BLIP2}, while mPLUG~\cite{mplug_owl} is a more recent method that suggests tuning both the large language model (LLM) and the visual encoder simultaneously. The comparison results are shown in Tab.~\ref{result_addi}, indicating that the visual tuning scheme adopted by LLaVa~\cite{LLaVa} generally performs the best. This is understandable because the primary task for layout generation is to adapt the input and output format, making the LLM the central component for tuning. Additionally, the existing layout data is still limited in both quality and quantity and aligning the visual encoder with such data would weaken its general feature extraction ability.

\subsection{Training Expenditure Analysis}
The utilization of LLMs brings better performance but also a potentially larger burden on computation. In this section, we present training-time experiments to demonstrate that the increased complexity of introducing LLMs is manageable in the layout generation task. The results are shown in Table~\ref{efficiency_test}. Interestingly, our method, despite incorporating a much larger model (LLaVa-7B or 13B), requires significantly fewer epochs to converge compared to previous methods (2 epochs vs. 300 epochs) and can thus save 50\% time (4,186 s vs. 9,030 s). This improvement is likely due to the spatial arranging knowledge implicitly encoded in the pre-trained LLM models and the superior instruction-following ability brought by instruction-tuning. Additionally, by using the zero3\_offload script for DeepSpeed, the LLM can be tuned on constrained GPU devices, such as 8 x NVIDIA A10 GPUs with only 24 GB of memory each, which is comparable with the requirement of previous methods. Furthermore, using the LoRA scheme can further reduce training time and memory requirements, making it a better alternative than full tuning when adopting larger models ($>$13B). In summary, using LLM for layout generation is promising for achieving both better effectiveness and improved efficiency.

\bibliographystyle{IEEEtran}
\bibliography{reference}
